\documentclass[sigconf,nonacm]{acmart}
\usepackage{threeparttable}
\usepackage{multirow}
\usepackage{tikz}
\usepackage{pgfplots}
\usepackage{pgfplotstable}
\usepackage{xcolor}
\usepackage{subcaption}
\usepackage{enumitem}
\usepackage{float}
\usepackage{stfloats}
\usepackage{diagbox}

\pgfplotsset{compat=1.18}
    \def\addlegendimage{\csname pgfplots@addlegendimage\endcsname}

\definecolor{mae}{HTML}{32012F}
\definecolor{rmse}{HTML}{F97300}

\AtBeginDocument{%
  \providecommand\BibTeX{{%
    \normalfont B\kern-0.5em{\scshape i\kern-0.25em b}\kern-0.8em\TeX}}}

\setcopyright{acmlicensed}
\copyrightyear{2018}
\acmYear{2018}
\acmDOI{XXXXXXX.XXXXXXX}

\acmConference[Conference acronym 'XX]{Make sure to enter the correct
  conference title from your rights confirmation emai}{June 03--05,
  2018}{Woodstock, NY}
\acmISBN{978-1-4503-XXXX-X/18/06}

\newtheorem{definition}{Definition}

\begin{document}

\title{PTrajM: Efficient and Semantic-rich Trajectory Learning with Pretrained Trajectory-Mamba}

\author{Yan Lin}
\authornote{Both authors contributed equally to this research.}
\email{lyan@cs.aau.dk}
\affiliation{%
  \institution{Aalborg University}
  \city{Aalborg}
  \country{Denmark}
}

\author{Yichen Liu}
\authornotemark[1]
\author{Zeyu Zhou}
\email{{liuyichen, zeyuzhou}@bjtu.edu.cn}
\affiliation{%
  \institution{Beijing Jiaotong University}
  \city{Beijing}
  \country{China}
}

\author{Haomin Wen}
\email{haominwe@andrew.cmu.edu}
\affiliation{%
  \institution{Carnegie Mellon University}
  \city{Pittsburgh}
  \country{USA}
}

\author{Erwen Zheng}
\email{zhengew@bjtu.edu.cn}
\affiliation{%
  \institution{Beijing Jiaotong University}
  \city{Beijing}
  \country{China}
}

\author{Shengnan Guo}
\author{Youfang Lin}
\email{{guoshn, yflin}@bjtu.edu.cn}
\affiliation{%
  \institution{Beijing Jiaotong University}
  \city{Beijing}
  \country{China}
}

\author{Huaiyu Wan}
\authornote{Corresponding author.}
\email{hywan@bjtu.edu.cn}
\affiliation{%
  \institution{Beijing Jiaotong University}
  \city{Beijing}
  \country{China}
}

\renewcommand{\shortauthors}{Yan Lin, Yichen Liu, et al.}

\begin{abstract}
Vehicle trajectories provide crucial movement information for various real-world applications. To better utilize vehicle trajectories, it is essential to develop a trajectory learning approach that can effectively and efficiently extract rich semantic information, including movement behavior and travel purposes, to support accurate downstream applications. However, creating such an approach presents two significant challenges. First, movement behavior are inherently spatio-temporally continuous, making them difficult to extract efficiently from irregular and discrete trajectory points. Second, travel purposes are related to the functionalities of areas and road segments traversed by vehicles. These functionalities are not available from the raw spatio-temporal trajectory features and are hard to extract directly from complex textual features associated with these areas and road segments.

To address these challenges, we propose PTrajM, a novel method capable of efficient and semantic-rich vehicle trajectory learning. To support efficient modeling of movement behavior, we introduce Trajectory-Mamba as the learnable model of PTrajM. By integrating movement behavior parameterization and a trajectory state-space model, Trajectory-Mamba effectively extracts continuous movement behavior while being more computationally efficient than existing structures. To facilitate efficient extraction of travel purposes, we propose a travel purpose-aware pre-training procedure. This aligns the learned trajectory embeddings of Trajectory-Mamba with the travel purposes identified by the road and POI encoders through contrastive learning. This way, PTrajM can discern the travel purposes of trajectories without additional computational resources during its embedding process. Extensive experiments on two real-world datasets and comparisons with several state-of-the-art trajectory learning methods demonstrate the effectiveness of PTrajM.
Code is available at \url{https://anonymous.4open.science/r/PTrajM-C973}.
\end{abstract}

\maketitle

\section{Introduction}
\label{sec:introduction}
Vehicle trajectories, which are sequences of (location, time) pairs, record the movement of vehicles during their journeys. With the widespread adoption of location recording devices, such as in-vehicle navigation systems and smartphones, these trajectories have become increasingly accessible. Concurrently, advancements in intelligent traffic systems have highlighted the value of vehicle trajectories in providing valuable movement information. Such information is crucial for various real-world applications, including movement prediction~\cite{DBLP:conf/ijcai/WuCSZW17,DBLP:journals/www/YanZSYD23}, anomaly detection~\cite{DBLP:conf/icde/Liu0CB20,DBLP:journals/pvldb/HanCMG22}, trajectory clustering~\cite{DBLP:conf/ijcnn/YaoZZHB17}, trajectory similarity measurement~\cite{DBLP:conf/kdd/YaoHDCHB22,DBLP:journals/tkde/HuCFFLG24}, and travel time estimation~\cite{DBLP:conf/sigmod/Yuan0BF20,DBLP:journals/pacmmod/LinWHGYLJ23}. The growing availability and use of vehicle trajectories drive the development of trajectory learning models, which facilitate the implementation of applications based on these trajectories.

To enhance the performance of real-world applications, it is important to develop a trajectory learning model that can effectively extract rich semantic information, including movement behavior and travel purposes, from these trajectories. Additionally, extracting this information efficiently is necessary to reduce computational burden and improve response time. However, achieving these goals is hampered by several challenges.

\begin{figure}
    \centering
    \includegraphics[width=1.0\linewidth]{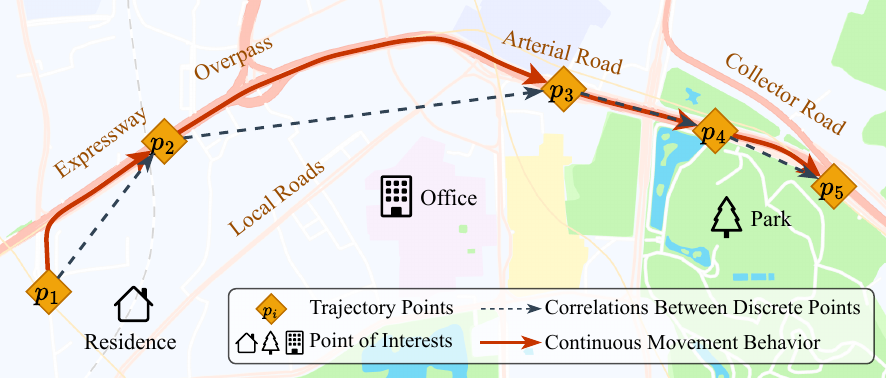}
    \caption{Motivation example of a vehicle trajectory.}
    \label{fig:motivation}
\end{figure}

\textbf{First, extracting continuous movement patterns from irregular and discrete trajectory points is challenging.} For instance, consider the vehicle trajectory $\mathcal T=\langle p_1, p_2, \dots, p_5 \rangle$ in Figure~\ref{fig:motivation}. The movement behavior of the vehicle is spatio-temporally continuous, represented by the red solid lines in the figure. However, this movement behavior is recorded by discrete trajectory points. The correlations between the points, illustrated by the grey dotted lines, do not accurately represent the continuous movement behavior. Common sequential models, including Recurrent Neural Networks (RNN)~\cite{hochreiter1997long,DBLP:journals/corr/ChungGCB14} and Transformers~\cite{DBLP:conf/nips/VaswaniSPUJGKP17}, only consider the correlations between discrete points and are thus ineffective at modeling movement behavior from vehicle trajectories.

Existing efforts~\cite{DBLP:conf/ijcai/LiangOYWTZ21,DBLP:conf/cikm/LiangOWLCZZZ22,10375102} have utilized Neural Ordinary Differential Equations (NeuralODE)~\cite{DBLP:conf/nips/ChenRBD18,DBLP:conf/nips/KidgerMFL20} or specially designed kernels coupled with Transformers to explicitly extract continuous movement behavior from discrete trajectory points. Despite the performance improvements, these solutions bring significant computational costs that hurt model efficiency. Specifically, NeuralODEs involve solving integral equations and are slow even on modern hardware, while Transformers involve quadratic computational complexity. Thus, efficiently modeling continuous movement patterns from discrete trajectory points remains an unsolved problem.

\textbf{Second, extracting travel purposes from the raw spatio-temporal features of trajectories or associated textual features is non-trivial.} Using the trajectory $\mathcal T$ in Figure~\ref{fig:motivation} as an example, the vehicle starts from a residential area, travels through an expressway and an arterial road, and finally reaches a park, indicating a travel purpose related to recreational or leisure activities. In other words, the functionalities of the traversed points of interest (POIs) and roads contain the travel purposes of vehicles. However, this information is not available from the raw spatio-temporal features of vehicle trajectories, and extracting it from the textual features associated with POIs and roads can be difficult.

The development of Language Models (LM)~\cite{DBLP:conf/naacl/DevlinCLT19,DBLP:conf/nips/BrownMRSKDNSSAA20,DBLP:conf/acl/DuQLDQY022} in recent years enables the extraction of complex functionalities of POIs and roads from their textual descriptions. This capability has been explored by some recent trajectory learning efforts~\cite{DBLP:journals/corr/abs-2405-12459}. However, incorporating LMs into a trajectory learning model brings a significant computational burden, as LMs are usually much larger in model size compared to standard trajectory learning models. Thus, efficiently incorporating POI and road functionalities to extract travel purposes remains an open question.

To address the aforementioned challenges, we propose \textit{\underline{P}retrained \underline{Traj}ectory-\underline{M}amba} (\textbf{PTrajM}), a novel method for efficient and semantically rich vehicle trajectory learning. PTrajM consists of two critical components: Trajectory-Mamba and travel purpose-aware pre-training, designed to achieve its objectives. The first component, Trajectory-Mamba, is a trajectory encoder that transforms vehicle trajectories into embedding vectors, facilitating the efficient extraction of continuous movement behavior. This model primarily comprises Traj-Mamba blocks, which incorporate movement behavior parameterization and trajectory state-space models (Traj-SSM) for effective movement behavior extraction. The second component, travel purpose-aware pre-training, enables the efficient modeling of travel purposes. It extracts travel purposes by utilizing road and POI encoders, mapping trajectories into road and POI views. It then aligns the mapped trajectory embeddings from Trajectory-Mamba with these views. After pre-training, PTrajM can discern travel purposes without requiring additional computational resources during the embedding process.

The primary contributions of the paper are summarized as follows:
\begin{itemize}[leftmargin=*]
    \item We propose a novel vehicle trajectory method called PTrajM, which efficiently extracts rich semantic information, including movement behavior and travel purposes, from vehicle trajectories.
    \item We design Trajectory-Mamba as the learnable component of PTrajM, which efficiently extracts movement behavior from vehicle trajectories and maps these trajectories into their embedding vectors.
    \item We introduce travel purpose-aware pre-training as the training procedure of PTrajM. It aligns the learned embeddings of Trajectory-Mamba with travel purposes, represented by road and POI views of vehicle trajectories, for efficient extraction of travel purposes.
    \item We conduct extensive experiments on two real-world vehicle trajectory datasets and compare various vehicle trajectory learning methods, demonstrating that PTrajM meets its design goals.
\end{itemize}

\section{Related Works} 
\label{sec:related-works}
Vehicle trajectory learning methods extract valuable information from vehicle trajectories to perform various tasks. These methods can be broadly categorized into end-to-end trajectory learning methods and pre-trained trajectory embeddings.

\subsection{End-to-end Trajectory Learning Methods}
End-to-end methods are tailored for specific tasks and are typically trained with task-specific labels. Trajectory prediction methods, such as DeepMove~\cite{DBLP:conf/www/FengLZSMGJ18}, HST-LSTM~\cite{DBLP:conf/ijcai/Kong018}, and ACN~\cite{DBLP:conf/wsdm/MiaoLZW20}, leverage Recurrent Neural Networks (RNN)~\cite{hochreiter1997long,DBLP:journals/corr/ChungGCB14} to capture sequential correlations in trajectories. PreCLN~\cite{DBLP:journals/www/YanZSYD23}, on the other hand, uses Transformers~\cite{DBLP:conf/nips/VaswaniSPUJGKP17} to process vehicle trajectories. Trajectory classification methods, including TrajFormer~\cite{DBLP:conf/cikm/LiangOWLCZZZ22}, MainTUL~\cite{DBLP:conf/ijcai/ChenLHYJD22}, and TULRN~\cite{DBLP:journals/www/SangXCZ23}, classify trajectories into their respective class labels, typically identifying user or driver IDs. Trajectory similarity measurement methods, such as NeuTraj~\cite{DBLP:conf/icde/YaoCZB19} and GTS~\cite{DBLP:journals/www/ZhouHYCZ23}, use learnable models to efficiently compute the similarities between trajectories.

End-to-end methods are straightforward to implement and have their advantages. However, these methods are tailored to specific tasks and cannot be easily repurposed for other tasks. This necessitates designing, training, and storing separate models for each task, which can impact computational resources and storage efficiency. Additionally, the effectiveness of end-to-end methods depends on the abundance of task-specific labels, which cannot always be guaranteed.

\subsection{Pre-trained Trajectory Embeddings}
To address the limitations of end-to-end methods, there is a growing interest in pre-training trajectory embeddings that can be utilized across various tasks. This approach involves learning trajectory encoders that map vehicle trajectories into embedding vectors, which can then be used with prediction modules. Among these methods, trajectory2vec~\cite{DBLP:conf/ijcnn/YaoZZHB17} uses an auto-encoding framework~\cite{hinton2006reducing} to compress each sequence into an embedding vector. t2vec~\cite{DBLP:conf/icde/LiZCJW18} employs a denoising auto-encoding framework to enhance its resilience to trajectory noise. Trembr~\cite{DBLP:journals/tist/FuL20} leverages auto-encoding techniques to effectively extract road network and temporal information embedded in trajectories. SML~\cite{DBLP:journals/kbs/ZhouDGWZ21} integrates contrastive learning~\cite{DBLP:journals/corr/abs-1807-03748} to learn embedding vectors for trajectories. START~\cite{DBLP:conf/icde/JiangPRJLW23} introduces a comprehensive approach to trajectory embedding learning by combining a masked language model~\cite{DBLP:conf/naacl/DevlinCLT19} with SimCLR~\cite{DBLP:conf/icml/ChenK0H20} to enhance its learning capability. MMTEC~\cite{10375102} utilizes maximum entropy coding~\cite{DBLP:conf/nips/LiuWLW22} to learn trajectory embeddings that perform consistently across different tasks.

Despite the promising progress made by existing efforts, as discussed in Section~\ref{sec:introduction}, there are still challenges in efficiently extracting rich semantic information from vehicle trajectories due to their complexity.

\section{Preliminaries} 
\label{sec:preliminaries}

\subsection{Definitions}
\begin{definition}
[Vehicle Trajectory]
A vehicle trajectory $\mathcal{T}$ is defined as a sequence of trajectory points: $\mathcal{T} = \langle p_1, p_2, \dots, p_n \rangle$, where $n$ is the number of points. Each point $p_i = (\mathrm{lng}_i, \mathrm{lat}_i, t_i)$ consists of the longitude $\mathrm{lng}_i$, latitude $\mathrm{lat}_i$, and timestamp $t_i$, representing the vehicle's location at a specific time.
\end{definition}

\begin{definition}
[Road Network]
A road network is modeled as a directed graph $\mathcal G=(\mathcal V, \mathcal E)$. Here, $\mathcal V$ is a set of nodes, with each node $v_i \in \mathcal V$ representing an intersection between road segments or the end of a segment. $\mathcal E$ is a set of edges, with each edge $e_i \in \mathcal E$ representing a road segment linking two nodes. An edge is defined by its starting and ending nodes, and a textual description including the name and type of the road: $e_i=(v_j, v_k, \mathrm{desc}_i^\mathrm{Road})$.
\end{definition}

\begin{definition}
[Point of Interest]
A point of interest (POI) is a location with specific cultural, environmental, or economic importance. We represent a POI as $l_i = (\mathrm{lng}_i, \mathrm{lat}_i, \mathrm{desc}_i^\mathrm{POI})$, where $\mathrm{lng}_i$ and $\mathrm{lat}_i$ are the coordinates of the POI, and $\mathrm{desc}_i$ is a textual description that includes the name, type, and address of the POI.
\end{definition}

\subsection{Problem Statement}
\textbf{Vehicle trajectory learning} aims to construct a learning model $f_\theta$, where $\theta$ is the set of learnable parameters. Given a vehicle trajectory $\mathcal{T}$, the model calculates its embedding vector as $\boldsymbol{e}_\mathcal{T} = f_\theta(\mathcal{T})$. This embedding vector $\boldsymbol{e}_\mathcal{T}$ captures the semantic information of $\mathcal{T}$ and can be used in subsequent applications by adding prediction modules.

\begin{figure*}
    \centering
    \begin{subfigure}[b]{0.45\linewidth}
        \centering
        \includegraphics[width=1.0\linewidth]{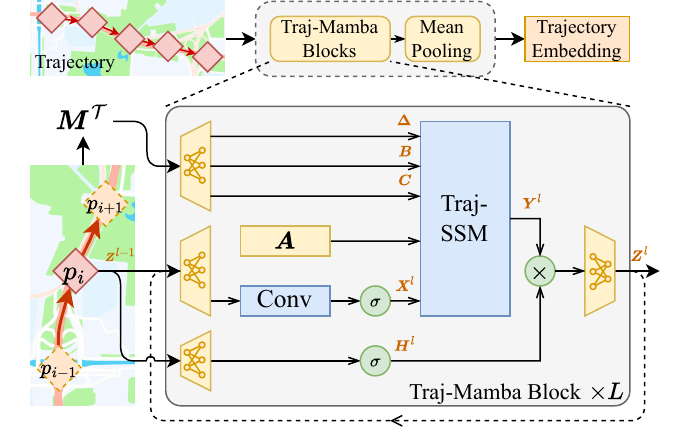}
        \caption{Structure of Trajectory-Mamba.}
        \label{fig:traj-mamba}
    \end{subfigure}
    \hfill
    \begin{subfigure}[b]{0.54\linewidth}
        \centering
        \includegraphics[width=1.0\linewidth]{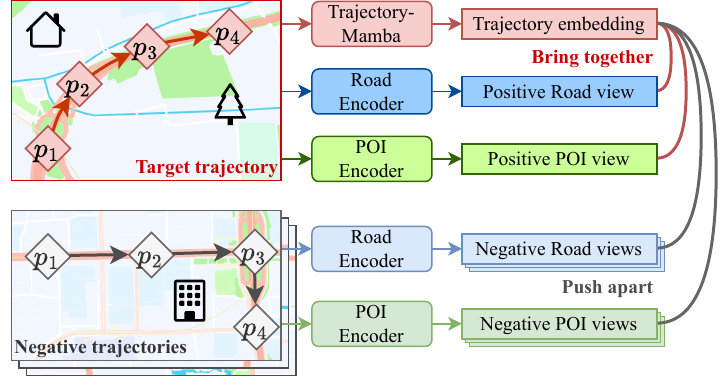}
        \caption{Travel purpose-aware pre-training.}
        \label{fig:pretrain}
    \end{subfigure}
\caption{The framework of PTrajM.}
\label{fig:framework}
\end{figure*}

\section{Methodology}
\subsection{Overview}
We propose a novel, efficient, and semantic-rich trajectory learning method named PTrajM. Figure~\ref{fig:framework} provides an overview of PTrajM's framework. It consists of two main components: the Trajectory-Mamba model for efficient trajectory embedding and the travel purpose-aware pre-training procedure that enables semantic-rich trajectory learning.

The Trajectory-Mamba model, illustrated in Figure~\ref{fig:traj-mamba}, is a trajectory encoder and the learnable model of PTrajM. It extracts rich semantic information from trajectories and incorporates this information into the embedding vectors of trajectories. Trajectory-Mamba primarily consists of Traj-Mamba blocks, which are inspired by the Mamba2 structure~\cite{DBLP:journals/corr/abs-2405-21060}. These blocks efficiently extract continuous movement behavior from trajectories using movement behavior parameterization and the trajectory state-space model (Traj-SSM).

The travel purpose-aware pre-training procedure, shown in Figure~\ref{fig:pretrain}, enhances Trajectory-Mamba's ability to extract travel purposes without adding significant computational overhead. This procedure aligns the learned trajectory embeddings of Trajectory-Mamba with the travel purposes identified by the road and POI encoders through contrastive learning. After pre-training, Trajectory-Mamba can extract rich semantic information from trajectories without adding more computational or storage resources to its embedding process.

The following sections provide detailed explanations of the two modules of PTrajM.

\subsection{Trajectory-Mamba}
To efficiently extract movement behavior, we propose the Trajectory-Mamba model. It consists mainly of Traj-Mamba blocks, which parameterize the movement behavior of trajectories and incorporate the trajectory state-space model (Traj-SSM) to model continuous movement behavior efficiently. Finally, the output from the Traj-Mamba blocks is fed into a mean pooling layer to obtain the embedding vectors of vehicle trajectories.

\subsubsection{Movement Behavior Feature Extraction}
We begin by extracting movement behavior from the raw features of vehicle trajectories into latent and higher-order features.

For each trajectory point $p_i$ in a trajectory $\mathcal{T}=\langle p_1, p_2, \dots, p_n \rangle$, we employ a linear transformation layer to map its spatial coordinates $(\mathrm{lng}_i, \mathrm{lat}_i)$ into an embedding space $\mathbb{R}^{E}$, where $E$ represents the embedding dimension. Additionally, we transform its timestamp $t_i$ into a vector $\boldsymbol t_i \in \mathbb{R}^4$ of four features: the day of the week, the hour of the day, the minute of the hour, and the time difference in minutes relative to $t_1$. These four features are encoded into four embedding vectors using learnable Fourier encoding layers~\cite{DBLP:conf/nips/TancikSMFRSRBN20}. The embedding vectors are concatenated and then mapped into the embedding space $\mathbb{R}^{E}$ through a linear transformation layer.
Finally, the latent vector of the point $\boldsymbol z_i$ is obtained by adding the spatial and temporal latent vectors, formulated as follows:
\begin{equation}
    \boldsymbol z_i = \mathrm{Linear}(\langle \mathrm{lng}_i, \mathrm{lat}_i \rangle) + \mathrm{Linear}(\mathrm{Cat}(\mathrm{Fourier}(\boldsymbol t_i))),
\end{equation}
where $\mathrm{Cat}$ denotes vector concatenation, and $\mathrm{Fourier}$ denotes the Fourier encoding layer.

By gathering the latent vector for each point in $\mathcal{T}$, we obtain its sequence of latent vectors:
\begin{equation}
    \boldsymbol Z^\mathcal T = \langle \boldsymbol z_1, \boldsymbol z_2, \dots, \boldsymbol z_n \rangle \in \mathbb{R}^{n\times E}
\label{eq:sequence-of-latent-vectors}
\end{equation}

To facilitate the modeling of continuous movement behavior, we also extract high-order features, including speed, acceleration, and movement angle, for each point. The speed $v_i$ of each point $p_i (i>1)$ is calculated as:
\begin{equation}
v_i = \mathrm{Dist}((\mathrm{lng}_i, \mathrm{lat}_i), (\mathrm{lng}_{i-1}, \mathrm{lat}_{i-1})) / (t_i-t_{i-1}),
\end{equation}
where $\mathrm{Dist}$ calculates the shortest distance between two locations on the Earth's surface.
The acceleration $\mathrm{acc}_i$ is calculated as:
\begin{equation}
\mathrm{acc}_i = (v_i-v_{i-1})/(t_i-t_{i-1})
\end{equation}
The movement angle $\theta_i$, specifically the angle clockwise from true north to the target direction, is calculated as:
\begin{equation}
\begin{split}
L =& \sin(\mathrm{lng}_i-\mathrm{lng}_{i-1})\cdot\cos(\mathrm{lat}_i) \\
R =& \cos(\mathrm{lat}_{i-1})\cdot\sin(\mathrm{lat}_i) - \\ &\sin(\mathrm{lat}_{i-1})\cdot\cos(\mathrm{lat}_i)\cdot\cos(\mathrm{lng}_i-\mathrm{lng}_{i-1}) \\
\theta_i =& \arctan(L/R)
\end{split}
\end{equation}

Finally, we apply min-max normalization on these three features and concatenate them into a vector denoted as $\boldsymbol m_i=(v_i,\mathrm{acc}_i,\theta_i)$, representing the high-order features describing the movement behavior at point $p_i$. Note that $\boldsymbol m_1$ is set to $\boldsymbol m_2$.

By calculating the high-order features for each point in $\mathcal T$, we obtain its sequence of high-order features:
\begin{equation}
    \boldsymbol M^\mathcal T = \langle \boldsymbol m_1, \boldsymbol m_2, \dots, \boldsymbol m_n \rangle \in \mathbb R^{n\times 3}
\label{eq:sequence-of-high-order-features}
\end{equation}

\subsubsection{Traj-Mamba Block}
We propose Traj-Mamba blocks as the core components of Trajectory-Mamba, designed to utilize the features extracted above and model continuous movement behavior through movement behavior parameterization.

The Traj-Mamba blocks are arranged in a multi-layer structure. The $l$-th Traj-Mamba block takes the sequence $\boldsymbol{Z}^{l-1}\in \mathbb R^{n\times E}$ of latent vectors as input. This input is first processed through linear projection layer, followed by a convolution layer and a SiLU activation:
\begin{equation}
    \boldsymbol X^l = \sigma(\mathrm{Conv}(\mathrm{Linear}(\boldsymbol{Z}^{l-1}))),
\label{eq:input-mapping}
\end{equation}
where $\boldsymbol X^l\in \mathbb{R}^{n\times D}$, and $D$ is the model dimension. Here, $\mathrm{Conv}$ denotes the 1D causal convolution, and $\sigma$ represents the SiLU function. Note that the input to the first Traj-Mamba block, $\boldsymbol Z^0$, is the sequence $\boldsymbol Z^\mathcal T$ calculated in Equation~\ref{eq:sequence-of-latent-vectors}.

Next, we implement the movement behavior parameterization. This involves computing parameter matrices with the sequence $\boldsymbol M^\mathcal T$ of high-order movement behavior features calculated in Equation~\ref{eq:sequence-of-high-order-features}, which are then used in the Traj-SSM module within each Traj-Mamba block. Formally, we calculate three parameter matrices $\boldsymbol B$, $\boldsymbol C$, $\boldsymbol \Delta$ as follows:
\begin{equation}
\begin{split}
\boldsymbol{B}, \boldsymbol{C}, \boldsymbol{\hat{\Delta}} &= \mathrm{Linear}(\boldsymbol M^\mathcal T) \\
\boldsymbol \Delta &= \tau(\boldsymbol{\hat{\Delta}} + \boldsymbol{b}_\Delta ),
\end{split}
\end{equation}
where $\tau$ denotes the Softplus activation function, and $\boldsymbol{b}_\Delta$ is bias parameter of $\boldsymbol \Delta$. $\boldsymbol B \in \mathbb{R}^{n\times N}$ and $\boldsymbol C \in \mathbb{R}^{n\times N}$ are the input and output mapping matrices in Traj-SSM, respectively, $N$ is the state dimension. By parameterizing them with high-order movement behavior features, the model can accurately control how changes in movement behavior affect the processing of input features and output embeddings. Additionally, $\boldsymbol \Delta \in \mathbb{R}^{n\times H}$ is the timescale matrix that controls the rate at which Traj-SSM evolves over the trajectory sequence, where $H$ is the number of heads. Thus, parameterizing $\boldsymbol \Delta$ is crucial for capturing continuous movement behavior from discrete and irregular trajectory points.
Another parameter, $\boldsymbol A\in \mathbb R^{H}$, serves as the hidden state mapping matrix and is randomly initialized and regarded learnable parameters of the Traj-Mamba block.

Following the implementation of state space models (SSMs) in Mamba~\cite{DBLP:journals/corr/abs-2312-00752}, we use the timescale matrix $\boldsymbol\Delta$ to discretize $\boldsymbol A$ and $\boldsymbol B$ into $\boldsymbol{\bar{A}}$ and $\boldsymbol{\bar{B}}$ as follows:
\begin{equation}
\begin{split}
\boldsymbol{\bar{A}}&=\langle\exp(\boldsymbol\Delta_1^\top \odot \boldsymbol A),  \exp(\boldsymbol\Delta_2^\top \odot \boldsymbol A), \dots, \exp(\boldsymbol\Delta_n^\top\odot \boldsymbol A)
\rangle\\ 
\boldsymbol{\bar{B}}&= \langle \boldsymbol\Delta_1^\top \boldsymbol B_1 , \boldsymbol\Delta_2^\top \boldsymbol B_2 , \dots, \boldsymbol\Delta_n^\top \boldsymbol B_n  \rangle,
\end{split}
\label{eq:discretize}
\end{equation}
where $\boldsymbol{\bar{A}} \in \mathbb{R}^{n\times H}$ is discretized via zero-order hold (ZOH) and $\boldsymbol{\bar{B}}\in \mathbb{R}^{n\times H\times N}$ via Euler discretization. $\odot$ denotes the Hadamard product, and $\boldsymbol\Delta_i, \boldsymbol B_i$ are the $i$-th row of $\boldsymbol\Delta$ and $\boldsymbol B$, respectively. By discretize $\boldsymbol A$ and $\boldsymbol B$ with the timescale matrix $\boldsymbol \Delta$, the model can represent continuous movement behavior more accurately.

Using the discretized matrices and the sequence $\boldsymbol X^l$ calculated in Equation~\ref{eq:input-mapping}, we implement the trajectory state space model (Traj-SSM), denoted as:
\begin{equation}
\boldsymbol{Y}^l = \mathrm{TrajSSM}(\bar{\boldsymbol A}, \bar{\boldsymbol B}, \boldsymbol C)(\boldsymbol{X}^l),
\end{equation}
where we followed the implementation of multi-input SSM in Mamba2~\cite{DBLP:journals/corr/abs-2405-21060}, creating $H$ heads by reshaping the input $\boldsymbol{X}^l \in \mathbb{R}^{n\times D}$ into $\boldsymbol X^l\in\mathbb{R}^{n \times H \times \frac{D}{H}}$, and aggregate the heads by reshaping the output $\boldsymbol Y^l\in\mathbb{R}^{n \times H \times \frac{D}{H}}$ back to $\boldsymbol{Y}^l \in \mathbb{R}^{n\times D}$.

Similar to RNNs, Traj-SSM can be computed in a recurrent form. For the $i$-th step and $j$-th head, the recurrence formulations of Traj-SSM when $D/H=1$ are as follows:
\begin{equation}
\begin{split}
\boldsymbol h_{ij} &= \bar{\boldsymbol A}_{ij}\boldsymbol I\boldsymbol h_{i-1,j} + \bar{\boldsymbol B}_{ij}\boldsymbol x_{ij} \\
\boldsymbol y_{ij} &= \boldsymbol C_i\boldsymbol h_{ij},
\end{split}
\label{eq:trajssm-recurrence-form}
\end{equation}
where $\boldsymbol h_{ij}\in \mathbb{R}^{N\times1}$ is the hidden state, $\boldsymbol I\in \mathbb R^{N\times N}$ is the identity matrix. $\boldsymbol x_{ij} \in \mathbb R,\boldsymbol y_{ij}\in\mathbb{R}$, $\bar{\boldsymbol A}_{ij} \in \mathbb R$, $\bar{\boldsymbol B}_{ij}\in\mathbb{R}^{N\times 1}$ and $\boldsymbol C_i\in\mathbb{R}^{1\times  N}$ are $\boldsymbol{X}^{l}[i,j,:]$,  $\boldsymbol{Y}^{l}[i,j,:]$, $\bar{\boldsymbol{A}}[i,j]$, $\bar{\boldsymbol B}[i,j,:]$, and $\boldsymbol{C}[i,:]$, respectively. 
The above equation can be generalized to $D/H > 1$ by treating the input to the $j$-th head of Traj-SSM, $\boldsymbol{X}^{l}[:,j,:]$, as $D/H$ independent sequences and applying the equation to each sequence.
Additionally, we implement the hardware-efficient algorithm provided Mamba2, ensuring linear computational complexity with respect to the trajectory length $n$.

Besides Traj-SSM, another branch of the Traj-Mamba block consists of a linear mapping followed by a SiLU activation:
\begin{equation}
    \boldsymbol{H}^l = \sigma(\mathrm{Linear}(\boldsymbol{Z}^{l-1}))
\end{equation}

Finally, we combine the output of the two branches, $\boldsymbol Y^l$ and $\boldsymbol H^l$, obtaining the output of the $l$-th Traj-Mamba block as follows:
\begin{equation}
\boldsymbol{Z}^l = \mathrm{Linear}(\mathrm{Norm}(\boldsymbol{Y}^{l} \odot \boldsymbol{H}^{l}),
\end{equation}
where $\boldsymbol Z^l$ has the same shape as $\boldsymbol Z^{l-1}$. The normalization layer $\mathrm{Norm}$ can be implemented by LayerNorm, GroupNorm, or RMSNorm. $\boldsymbol Z^l$ can then be regarded as the input sequence to the $(l+1)$-th Traj-Mamba block.

\subsubsection{Trajectory Embedding}
We implement the Trajectory-Mamba model by stacking $L$ layers of Traj-Mamba blocks. Note that different layers do not share learnable parameters. Given with the sequence $\boldsymbol Z^\mathcal T$ from Equation~\ref{eq:sequence-of-latent-vectors} and the sequence $\boldsymbol M^\mathcal T$ from Equation~\ref{eq:sequence-of-high-order-features}, we take the output sequence from the final Traj-Mamba block and apply a mean pooling layer to derive the embedding vector $\boldsymbol z_\mathcal T \in \mathbb R^E$ of $\mathcal T$. For $L=2$, this process is expressed as:
\begin{equation}
    \boldsymbol z_\mathcal T = \mathrm{MeanPool}(
    \mathrm{TrajMamba}(\mathrm{TrajMamba}(\boldsymbol Z^\mathcal T, \boldsymbol M^\mathcal T), \boldsymbol M^\mathcal T)
    )
\end{equation}

By leveraging the linear time complexity of the Traj-SSM module and the ability to model continuous movement behavior through movement behavior parameterization, we achieve efficient and effective computation of trajectory embedding vectors.

\subsection{Travel Purpose-aware Pre-training}
To extract travel purposes without adding extra computational load to Trajectory-Mamba, we propose pre-training it using a travel purpose-aware scheme. First, we extract travel purposes from vehicle trajectories by modeling the textual information of their traversed roads and POIs using road and POI encoders. These encoders integrate the information into road and POI views. Then, we align Trajectory-Mamba's output embeddings with these views through contrastive learning.

\subsubsection{Road and POI views}
As analyzed in Section~\ref{sec:introduction}, the travel purpose of a vehicle trajectory is closely linked to the functions of the roads and POIs it traverses. Therefore, we propose to integrate travel purpose information of a trajectory into its road and POI views.

Given a trajectory $\mathcal T=\langle p_1, p_2, \dots, p_n \rangle$, we apply map-matching algorithms~\cite{DBLP:conf/adc/ChaoXH020} to map each of its points onto the road network $\mathcal G$ and obtain a map-matched trajectory composed of road segments: $\mathcal T^\mathrm{mm}=\langle e_1, e_2, \dots, e_n \rangle$, where $e_i$ is the road segment corresponding to point $p_i$. We then compute an embedding vector $\boldsymbol z_{e_i}\in \mathbb R^E$ for each road segment $e_i$ as follows:
\begin{equation}
\begin{split}
    \boldsymbol z_{e_i} = &\mathrm{Linear}(\mathrm{RoadIndexEmbed}(e_i)) + \\
    &\mathrm{Linear}(\mathrm{TextEmbed}(\mathrm{desc}^\mathrm{Road}_i)),
\end{split}
\label{eq:road-segment-embedding-vector}
\end{equation}
where $\mathrm{RoadIndexEmbed}$ is an index-fetch embedding layer that maps each unique road segment into a learnable embedding vector. $\mathrm{TextEmbed}$ is a pre-trained textual embedding module for mapping a line of text into an embedding vector, for which we use the \textit{text-embedding-3-large} model provided by OpenAI\footnote{\url{https://platform.openai.com/docs/guides/embeddings}}. 
Finally, we derive the road view $\boldsymbol z_{\mathcal T}^\mathrm{Road}\in \mathbb R^E$ of $\mathcal T$ by processing the sequence of embedding vectors of the road segments in $\mathcal T^\mathrm{mm}$ as follows:
\begin{equation}
    \boldsymbol z_{\mathcal T}^\mathrm{Road} = \mathrm{MeanPool}(\mathrm{RoadEnc}(\langle \boldsymbol z_{e_1}, \boldsymbol z_{e_2}, \dots, \boldsymbol z_{e_n}\rangle)),
\label{eq:road-view}
\end{equation}
where $\mathrm{RoadEnc}$ is the road encoder, which we implement using a 2-layer Transformer encoder.

To calculate the POI view, we first identify the closest POI $l_i$ to each trajectory point $p_i$ based on their geographical distance. Similar to Equation~\ref{eq:road-segment-embedding-vector}, we calculate an embedding vector $\boldsymbol z_{l_i} \in \mathbb R^E$ for each POI $l_i$ as follows:
\begin{equation}
\begin{split}
    \boldsymbol z_{l_i} = &\mathrm{Linear}(\mathrm{POIIndexEmbed}(l_i)) + \\
    &\mathrm{Linear}(\mathrm{TextEmbed}(\mathrm{desc}^\mathrm{POI}_i)),
\end{split}
\end{equation}
where $\mathrm{POIIndexEmbed}$ is another index-fetch embedding layer that assigns a learnable embedding vector for each unique POI, $\mathrm{TextEmbed}$ is the same textual embedding module used in Equation~\ref{eq:road-segment-embedding-vector}.
Finally, we calculate the POI view $\boldsymbol z_{\mathcal T}^\mathrm{POI} \in \mathbb R^E$ of $\mathcal T$ by processing the sequence of embedding vectors of the POIs correlated with $\mathcal T$ as follows:
\begin{equation}
    \boldsymbol z_{\mathcal T}^\mathrm{POI} = \mathrm{MeanPool}(\mathrm{POIEnc}(\langle \boldsymbol z_{l_1}, \boldsymbol z_{l_2}, \dots, \boldsymbol z_{l_n}\rangle)),
\label{eq:poi-view}
\end{equation}
where $\mathrm{POIEnc}$ is the POI encoder that we implement using another 2-layer Transformer encoder.

\subsubsection{Contrastive Learning}
After representing travel purposes as road and POI views, we align the output embedding vectors from Trajectory-Mamba with these two views using contrastive learning.

Given a batch of trajectories $\mathbb{T} = \{\mathcal{T}_1, \mathcal{T}_2, \dots, \mathcal{T}_B\}$ with a batch size of $B$, we can obtain their embedding vectors from Trajectory-Mamba as $\{\boldsymbol{z}_{\mathcal{T}_1}, \boldsymbol{z}_{\mathcal{T}_2}, \dots, \boldsymbol{z}_{\mathcal{T}_B}\}$. Their road views, as per Equation~\ref{eq:road-view}, are $\{\boldsymbol{z}_{\mathcal{T}_1}^\mathrm{Road}, \boldsymbol{z}_{\mathcal{T}_2}^\mathrm{Road}, \dots, \boldsymbol{z}_{\mathcal{T}_B}^\mathrm{Road}\}$. Their POI views, as per Equation~\ref{eq:poi-view}, are $\{\boldsymbol{z}_{\mathcal{T}_1}^\mathrm{POI}, \boldsymbol{z}_{\mathcal{T}_2}^\mathrm{POI}, \dots, \boldsymbol{z}_{\mathcal{T}_B}^\mathrm{POI}\}$. The similarity between $\mathcal{T}_i$ and the road and POI views of $\mathcal{T}_j$ is then calculated using the dot product as follows:
\begin{equation}
\begin{split}
    s_{ij}^\mathrm{Road} &= \boldsymbol{z}_{\mathcal{T}_i} \cdot \boldsymbol{z}_{\mathcal{T}_j}^\mathrm{Road} \\
    s_{ij}^\mathrm{POI} &= \boldsymbol{z}_{\mathcal{T}_i} \cdot \boldsymbol{z}_{\mathcal{T}_j}^\mathrm{POI} 
\end{split}
\end{equation}
Next, we apply the InfoNCE loss~\cite{DBLP:journals/corr/abs-1807-03748} to these similarities as follows:
\begin{equation}
\begin{split}
    \mathcal{L}_{\mathbb{T}}^\mathrm{Road} &= -\frac{1}{B} \sum_{i=1}^B \log \frac{\exp(s_{ii}^\mathrm{Road}/\tau)}{\sum_{j=1}^B \exp(s_{ij}^\mathrm{Road}/\tau)} \\
    \mathcal{L}_{\mathbb{T}}^\mathrm{POI} &= -\frac{1}{B} \sum_{i=1}^B \log \frac{\exp(s_{ii}^\mathrm{POI}/\tau)}{\sum_{j=1}^B \exp(s_{ij}^\mathrm{POI}/\tau)},
\end{split}
\end{equation}
where $\tau$ is the temperature parameter directly optimized during training as a log-parameterized multiplicative scalar~\cite{DBLP:conf/icml/RadfordKHRGASAM21}. $\mathcal{L}_{\mathbb{T}}^\mathrm{Road}$ and $\mathcal{L}_{\mathbb{T}}^\mathrm{POI}$ can be seen as maximizing the similarities between the trajectory embedding and the two views of the same trajectory in $\mathbb{T}$, while minimizing those between different trajectories. Finally, the contrastive learning loss is a combination of the two losses above:
\begin{equation}
    \mathcal{L}_{\mathbb{T}} = \frac{1}{2}(\mathcal{L}_{\mathbb{T}}^\mathrm{Road} + \mathcal{L}_{\mathbb{T}}^\mathrm{POI})
\end{equation}

After pre-training, the embedding vector from Trajectory-Mamba is aligned with the travel purposes represented by road and POI views. Additionally, the pre-training process does not add extra computational requirements to Trajectory-Mamba during its embedding process, thus maintaining its efficiency.

\section{Experiments}
We assess the effectiveness of PTrajM using two real-world vehicle trajectory datasets and compare its performance against several state-of-the-art methods.

\subsection{Datasets}
The two vehicle trajectory datasets are referred to as \textbf{Chengdu} and \textbf{Xian}. They consist of vehicle trajectories recorded by taxis operating in Chengdu and Xian, China, and were released by Didi\footnote{\url{https://gaia.didichuxing.com/}}. Due to the original trajectories having very dense sampling intervals, we retain a portion of the trajectory points through a three-hop resampling process, making most trajectories having sampling intervals of no less than 6 seconds. After resampling, trajectories with fewer than 5 or more than 120 trajectory points are considered anomalies and excluded.
Additionally, we retrieve the information of POIs within these datasets' areas of interest from the AMap API\footnote{\url{https://lbs.amap.com/api/javascript-api-v2}}, and obtain the road network topology and information from OpenStreetMap\footnote{\url{https://www.openstreetmap.org/}}.
The statistics of these datasets after the above preprocessing are listed in Table~\ref{table:dataset}.

\begin{table}[h]
    \centering
    \caption{Dataset statistics.}
    \begin{tabular}{ccc}
    \toprule
    Dataset & Chengdu & Xian \\
    \midrule
    Time span & 09/30 - 10/10, 2018 & 09/29 - 10/15, 2018 \\
    \#Trajectories & 140,000 & 210,000 \\
    \#Points & 18,832,411 & 18,267,440 \\
    \#Road segments & 4,315 & 3,392 \\
    \#POIs & 12,439 & 3,900 \\
    \bottomrule
\end{tabular}

    \label{table:dataset}
\end{table}

\subsection{Comparison Methods}
We include several state-of-the-art vehicle trajectory learning methods for comparison.

\begin{itemize}[leftmargin=*]
\item \textbf{t2vec}~\cite{DBLP:conf/kdd/Fang0ZHCGJ22}: Pre-trains the model by reconstructing original trajectories from low-sampling ones using a denoising auto-encoder.
\item \textbf{Trembr}~\cite{DBLP:journals/tist/FuL20}: Constructs an RNN-based seq2seq model to recover road segments and the time of the input trajectories.
\item \textbf{CTLE}~\cite{DBLP:conf/aaai/LinW0L21}: Pre-trains a bi-directional Transformer with two MLM tasks for location and hour predictions. The trajectory representation is obtained by applying mean pooling on point embeddings.
\item \textbf{Toast}~\cite{DBLP:conf/cikm/ChenLCBLLCE21}: Uses a context-aware node2vec model to generate segment representations and trains the model with an MLM-based task and a sequence discrimination task.
\item \textbf{TrajCL}~\cite{DBLP:conf/icde/Chang0LT23}: Introduces a dual-feature self-attention-based encoder and trains the model in a contrastive style using the InfoNCE loss.
\item \textbf{LightPath}~\cite{DBLP:conf/kdd/YangHGYJ23}: Constructs a sparse path encoder and trains it with a path reconstruction task and a cross-view contrastive task.
\item \textbf{START}~\cite{DBLP:conf/icde/JiangPRJLW23}: Includes a time-aware trajectory encoder and a GAT that considers the transitions between road segments. The model is trained with both an MLM task and a contrastive task based on SimCLR loss.
\end{itemize}

\subsection{Downstream Tasks}
\label{sec:downstream-tasks}
To assess the effectiveness of trajectory embeddings learned from PTrajM and comparison methods, we apply these embeddings to three representative downstream tasks.

\subsubsection{Destination Prediction}
This task involves predicting the destination of a trajectory. When calculating a trajectory $\mathcal{T}$'s embedding $\boldsymbol{z}_{\mathcal{T}}$, the last 5 points of $\mathcal{T}$ are omitted. A fully connected network then uses this embedding to predict the destination's coordinates. Mean Squared Error (MSE) is used to supervise the prediction by comparing the predicted and ground truth coordinates. Mean Absolute Error (MAE) and Root Mean Squared Error (RMSE) of the shortest distance on the Earth's surface serve as evaluation metrics.

\subsubsection{Arrival Time Estimation}
This task aims to predict the arrival time of a trajectory. Similar to the destination prediction task, the embedding vector of a trajectory $\mathcal{T}$ is calculated by omitting its last 5 points, and a fully connected network is used to predict the travel time. MSE supervises the prediction, while MAE, RMSE, and Mean Absolute Percentage Error (MAPE) are used as evaluation metrics.

\subsubsection{Similar Trajectory Search}
This task aims to identify the most similar trajectory to a query trajectory from a batch of candidates. Similarities between trajectories are calculated with the cosine similarity between their embeddings. Accuracy@N (Acc@1, Acc@5) and Mean Rank are used as evaluation metrics. 
Since most datasets don't have labeled data for this task, we create labels in the following way. We randomly select 1,000 trajectories from the test dataset. For each trajectory $\mathcal{T}$, we collect the odd-numbered points to form the query $\mathcal{T}^q$ and the even-numbered points to create the target $\mathcal{T}^t$. For each query, we discard the top 10 trajectories closest to the query, then randomly choose 5,000 additional trajectories from the rest of the test dataset to use as the database. To determine the distances between the query and other trajectories, we follow~\cite{DBLP:conf/kdd/Fang0ZHCGJ22}, downsampling them to a uniform length and computing the mean square error.

For the similar trajectory search task, the parameters of trajectory learning methods are fixed after pre-training. For the other two tasks, we can either fine-tune their parameters using task supervision or fix their parameters and only update the predictors' parameters. In the experiments, we denote the latter setting as \textit{without fine-tune} (\textbf{w/o ft}).

\begin{table}
    \caption{Overall performance of methods on destination prediction.}
    \label{tab:overall-destination-prediction}
    \resizebox{1.0\linewidth}{!}{
\begin{threeparttable}
\begin{tabular}{c|cc|cc}
\toprule
Dataset & \multicolumn{2}{c|}{Chengdu} & \multicolumn{2}{c}{Xian} \\
\midrule
\multirow{2}{*}{\diagbox[]{Method}{Metric}} & RMSE $\downarrow$ & MAE $\downarrow$ & RMSE $\downarrow$ & MAE $\downarrow$ \\
& (meters) & (meters) & (meters) & (meters) \\
\midrule
t2vec (w/o ft) & 2329.63$\pm$21.09& 1868.49$\pm$19.49 & 2582.14$\pm$46.79& 2235.27$\pm$39.44\\
Trembr (w/o ft) & 1787.18$\pm$92.01& 1419.58$\pm$88.95 & 2067.80$\pm$196.30& 1749.76$\pm$178.82\\
CTLE (w/o ft) & 3421.09$\pm$17.10 & 3041.49$\pm$23.49 & 3548.88$\pm$4.27& 3320.46$\pm$1.12\\
Toast (w/o ft) & 3434.84$\pm$9.55& 3061.91$\pm$14.99 & 3549.65$\pm$6.42& 3325.48$\pm$8.21\\
TrajCL (w/o ft) & \underline{1059.81$\pm$16.22} & \underline{865.48$\pm$10.60} & \underline{1268.41$\pm$19.57} & \underline{1054.21$\pm$18.54} \\
LightPath (w/o ft) & 2365.87$\pm$57.52 & 1948.97$\pm$57.78 & 2177.37$\pm$60.03& 1859.35$\pm$48.50\\
START (w/o ft) & 1347.13$\pm$30.72 & 1111.77$\pm$29.11 & 1406.06$\pm$18.42& 1173.62$\pm$17.18\\
\textbf{PTrajM} (w/o ft) & \textbf{332.06$\pm$7.20} & \textbf{260.38$\pm$6.75} & \textbf{470.54$\pm$8.56} & \textbf{365.62$\pm$6.24} \\
\midrule
t2vec & 579.30$\pm$11.94& 387.50$\pm$4.03& 482.64$\pm$2.67& 310.08$\pm$3.00\\
Trembr & 505.62$\pm$4.57& 376.88$\pm$7.34& 473.97$\pm$1.24& 301.45$\pm$4.98\\
CTLE & 430.19$\pm$52.65& 382.82$\pm$52.88& 477.70$\pm$48.25& 384.08$\pm$53.18\\
Toast & 480.52$\pm$82.39& 412.58$\pm$72.32& 523.76$\pm$67.04& 443.99$\pm$60.41\\
TrajCL & 365.50$\pm$19.14& 272.63$\pm$25.32& 383.39$\pm$7.30& 262.20$\pm$10.68\\
LightPath & 553.27$\pm$42.26& 360.86$\pm$56.41& 598.20$\pm$15.57& 348.61$\pm$19.32\\
START & \underline{333.10$\pm$10.47} & \underline{240.40$\pm$15.10} & \underline{319.00$\pm$4.27} & \underline{208.35$\pm$7.30} \\
\textbf{PTrajM} & \textbf{161.28$\pm$4.32} & \textbf{118.84$\pm$5.10} & \textbf{263.16$\pm$4.28} & \textbf{182.39$\pm$2.65} \\
\bottomrule
\end{tabular}
\begin{tablenotes}\footnotesize
\item[]{
    \textbf{Bold} denotes the best result, and \underline{underline} denotes the second-best result.
    $\downarrow$ means lower is better.
}
\end{tablenotes}
\end{threeparttable}
}
\end{table}

\subsection{Settings}
For both datasets, we split the trajectories into training, validation, and testing sets in an 8:1:1 ratio, with departure times in chronological order. PTrajM is pre-trained for 30 epochs on the training set, and downstream tasks are early-stopped based on the validation set. Final metrics are calculated using the testing set.

PTrajM is implemented using PyTorch~\cite{DBLP:conf/nips/PaszkeGMLBCKLGA19}. The six key hyperparameters and their optimal values are $B=128$, $L=4$, $N=128$, $H=4$, $E=256$, and $D=256$. We select parameters based on the MAE of the destination prediction task on Chengdu's validation set. The effectiveness of these parameters is reported in the next section. For model training, we use the Adam optimizer with an initial learning rate of 0.001. The experiments are conducted on servers equipped with Intel(R) Xeon(R) W-2155 CPUs and nVidia(R) TITAN RTX GPUs. Each set of experiments is run 5 times, and we report the mean and standard deviation of the metrics.

\begin{figure}
    \centering
    \pgfplotstableread[row sep=\\,col sep=&]{
v & mae & rmse \\
32 & 312.31 & 389.65 \\
64 & 277.42 & 349.42 \\
128 & 260.38 & 332.06 \\
256 & 263.84 & 333.90 \\
512 & 342.60 & 419.13 \\
}\BatchSize

\pgfplotstableread[row sep=\\,col sep=&]{
v & mae & rmse \\
1 & 311.31 & 397.61 \\
2 & 275.98 & 337.19 \\
3 & 268.23 & 335.75 \\
4 & 260.38 & 332.06 \\
5 & 271.54 & 339.37 \\
6 & 296.61 & 365.75 \\
}\NumLayers

\pgfplotstableread[row sep=\\,col sep=&]{
v & mae & rmse \\
16& 273.28 & 340.85 \\
32 & 267.02 & 333.34 \\
64 & 281.92 & 352.00 \\
128 & 260.38 & 332.06 \\
256 & 283.64 & 354.12 \\
}\StateDimN

\pgfplotstableread[row sep=\\,col sep=&]{
v & mae & rmse \\
2 & 262.58 & 340.55 \\
4 & 260.38 & 332.06 \\
8 & 261.17 & 330.28 \\
16 & 287.90 & 357.48 \\
32 & 264.18 & 328.94 \\
64 & 263.70 & 325.57 \\
}\NumHeadsH

\pgfplotstableread[row sep=\\,col sep=&]{
v & mae & rmse \\
32 & 400.40 & 493.55 \\
64 & 363.83 & 449.22 \\
128 & 293.31 & 368.65 \\
256 & 260.38 & 332.06 \\
512 & 279.00 & 349.75 \\
}\EmbedDimE

\pgfplotstableread[row sep=\\,col sep=&]{
v & mae & rmse \\
64 & 291.92 & 368.21 \\
128 & 296.21 & 369.28 \\
256 & 260.38 & 332.06 \\
512 & 260.89 & 321.65 \\
}\ModelDimD

\newcommand{\maeMin}{252}
\newcommand{\maeMax}{420}
\newcommand{\maeTick}{20}

\newcommand{\rmseMin}{290}
\newcommand{\rmseMax}{500}
\newcommand{\rmseTick}{25}

\tikzset{every plot/.style={line width=1.5pt}}

\ref{fig:shared-legend}

\begin{subfigure}[b]{0.48\linewidth}
    \begin{tikzpicture}
    \begin{axis}[
        width=1.15\linewidth, height=1.0\linewidth,
        yticklabel style = {mae},
        ymin=\maeMin, ymax=\maeMax,
        ytick distance={\maeTick},
        xtick=data,
        symbolic x coords={32, 64, 128, 256, 512},
        legend entries={MSE (meters), RMSE (meters)},
        legend to name=fig:shared-legend,
        legend columns=-1,
    ]
    \addplot[color=mae,mark=square*] table[x=v,y=mae]{\BatchSize};
    \addlegendentry{MSE (meters)}
    \addlegendimage{/pgfplots/refstyle=rmseStyle}
    \addlegendentry{RMSE (meters)}
    \end{axis}
    \begin{axis}[
        width=1.15\linewidth, height=1.0\linewidth,
        yticklabel style = {rmse},
        axis y line*=right,
        axis x line=none, 
        ylabel near ticks,
        ymin=\rmseMin, ymax=\rmseMax,
        ytick distance={\rmseTick},
        xtick=data, 
        symbolic x coords={32, 64, 128, 256, 512},
    ]
    \addplot[color=rmse,mark=triangle*] table[x=v,y=rmse]{\BatchSize};
    \label{rmseStyle}
    \end{axis}
    \end{tikzpicture}
    \caption{Batch Size $B$}
    \label{fig:batch-size}
\end{subfigure}
\hfill
\begin{subfigure}[b]{0.48\linewidth}
    \begin{tikzpicture}
    \begin{axis}[
        width=1.15\linewidth, height=1.0\linewidth,
        yticklabel style = {mae},
        ymin=\maeMin, ymax=\maeMax,
        ytick distance={\maeTick},
        xtick=data,
        symbolic x coords={1, 2, 3, 4, 5, 6},
    ]
    \addplot[color=mae,mark=square*] table[x=v,y=mae]{\NumLayers};
    \end{axis}
    \begin{axis}[
        width=1.15\linewidth, height=1.0\linewidth,
        yticklabel style = {rmse},
        axis y line*=right,
        axis x line=none, 
        ylabel near ticks,
        ymin=\rmseMin, ymax=\rmseMax,
        ytick distance={\rmseTick},
        xtick=data, 
        symbolic x coords={1, 2, 3, 4, 5, 6},
    ]
    \addplot[color=rmse,mark=triangle*] table[x=v,y=rmse]{\NumLayers};
    \end{axis}
    \end{tikzpicture}
    \caption{Number of Layers $L$}
    \label{fig:num-layers}
\end{subfigure}

\begin{subfigure}[b]{0.48\linewidth}
    \begin{tikzpicture}
    \begin{axis}[
        width=1.15\linewidth, height=1.0\linewidth,
        yticklabel style = {mae},
        ymin=\maeMin, ymax=\maeMax,
        ytick distance={\maeTick},
        xtick=data,
        symbolic x coords={16,32,64,128,256},
    ]
    \addplot[color=mae,mark=square*] table[x=v,y=mae]{\StateDimN};
    \end{axis}
    \begin{axis}[
        width=1.15\linewidth, height=1.0\linewidth,
        yticklabel style = {rmse},
        axis y line*=right,
        axis x line=none, 
        ylabel near ticks,
        ymin=\rmseMin, ymax=\rmseMax,
        ytick distance={\rmseTick},
        xtick=data, 
        symbolic x coords={16,32,64,128,256},
    ]
    \addplot[color=rmse,mark=triangle*] table[x=v,y=rmse]{\StateDimN};
    \end{axis}
    \end{tikzpicture}
    \caption{State Dimension $N$}
    \label{fig:state-dim-n}
\end{subfigure}
\hfill
\begin{subfigure}[b]{0.48\linewidth}
    \begin{tikzpicture}
    \begin{axis}[
        width=1.15\linewidth, height=1.0\linewidth,
        yticklabel style = {mae},
        ymin=\maeMin, ymax=\maeMax,
        ytick distance={\maeTick},
        xtick=data,
        symbolic x coords={2,4,8,16,32,64},
    ]
    \addplot[color=mae,mark=square*] table[x=v,y=mae]{\NumHeadsH};
    \end{axis}
    \begin{axis}[
        width=1.15\linewidth, height=1.0\linewidth,
        yticklabel style = {rmse},
        axis y line*=right,
        axis x line=none, 
        ylabel near ticks,
        ymin=\rmseMin, ymax=\rmseMax,
        ytick distance={\rmseTick},
        xtick=data, 
        symbolic x coords={2,4,8,16,32,64},
    ]
    \addplot[color=rmse,mark=triangle*] table[x=v,y=rmse]{\NumHeadsH};
    \end{axis}
    \end{tikzpicture}
    \caption{Number of Heads $H$}
    \label{fig:num-heads-h}
\end{subfigure}

\begin{subfigure}[b]{0.48\linewidth}
    \begin{tikzpicture}
    \begin{axis}[
        width=1.15\linewidth, height=1.0\linewidth,
        yticklabel style = {mae},
        ymin=\maeMin, ymax=\maeMax,
        ytick distance={\maeTick},
        xtick=data,
        symbolic x coords={32,64,128,256,512},
    ]
    \addplot[color=mae,mark=square*] table[x=v,y=mae]{\EmbedDimE};
    \end{axis}
    \begin{axis}[
        width=1.15\linewidth, height=1.0\linewidth,
        yticklabel style = {rmse},
        axis y line*=right,
        axis x line=none, 
        ylabel near ticks,
        ymin=\rmseMin, ymax=\rmseMax,
        ytick distance={\rmseTick},
        xtick=data, 
        symbolic x coords={32,64,128,256,512},
    ]
    \addplot[color=rmse,mark=triangle*] table[x=v,y=rmse]{\EmbedDimE};
    \end{axis}
    \end{tikzpicture}
    \caption{Embed Dimension $E$}
    \label{fig:embed-dim-e}
\end{subfigure}
\hfill
\begin{subfigure}[b]{0.48\linewidth}
    \begin{tikzpicture}
    \begin{axis}[
        width=1.15\linewidth, height=1.0\linewidth,
        yticklabel style = {mae},
        ymin=\maeMin, ymax=\maeMax,
        ytick distance={\maeTick},
        xtick=data,
        symbolic x coords={64,128,256,512},
    ]
    \addplot[color=mae,mark=square*] table[x=v,y=mae]{\ModelDimD};
    \end{axis}
    \begin{axis}[
        width=1.15\linewidth, height=1.0\linewidth,
        yticklabel style = {rmse},
        axis y line*=right,
        axis x line=none, 
        ylabel near ticks,
        ymin=\rmseMin, ymax=\rmseMax,
        ytick distance={\rmseTick},
        xtick=data, 
        symbolic x coords={64,128,256,512},
    ]
    \addplot[color=rmse,mark=triangle*] table[x=v,y=rmse]{\ModelDimD};
    \end{axis}
    \end{tikzpicture}
    \caption{Model Dimension $D$}
    \label{fig:model-dim-d}
\end{subfigure}
    \caption{Effectiveness of hyper-parameters.}
    \label{fig:hyper-parameter}
\end{figure}
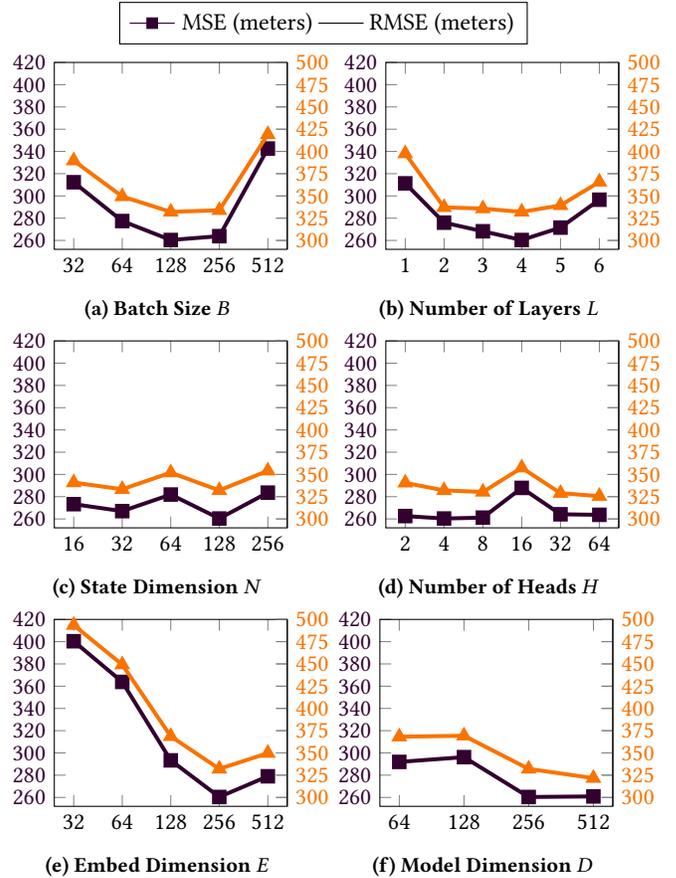

\begin{table*}
    \caption{Overall performance of methods on arrival time estimation.}
    \label{tab:overall-arrival-time-estimation}
\begin{threeparttable}
\begin{tabular}{c|ccc|ccc}
\toprule
Dataset & \multicolumn{3}{c|}{Chengdu} & \multicolumn{3}{c}{Xian} \\
\midrule
\multirow{2}{*}{\diagbox[]{Method}{Metric}} & RMSE $\downarrow$ & MAE $\downarrow$ & MAPE $\downarrow$ & RMSE $\downarrow$ & MAE $\downarrow$ & MAPE $\downarrow$ \\
& (seconds) & (seconds) & (\%) & (seconds) & (seconds) & (\%) \\
\midrule
t2vec (w/o ft) & 138.30$\pm$1.63& 79.74$\pm$1.98& 18.71$\pm$0.57& 207.11$\pm$4.12 & 117.86$\pm$4.74 & 16.01$\pm$0.52 \\
Trembr (w/o ft) & 159.60$\pm$8.40& 110.36$\pm$6.51& 29.50$\pm$1.00& 435.04$\pm$5.17 & 337.35$\pm$3.66 & 47.11$\pm$0.06 \\
CTLE (w/o ft) & 135.59$\pm$4.68& 63.45$\pm$5.08& 13.99$\pm$1.27& 272.88$\pm$60.42 & 176.16$\pm$72.08 & 31.84$\pm$10.46 \\
Toast (w/o ft) & 149.67$\pm$8.22& 79.69$\pm$9.59& 17.89$\pm$0.82& 299.94$\pm$51.68 & 205.49$\pm$54.15 & 32.55$\pm$5.71 \\
TrajCL (w/o ft) & 136.56$\pm$3.90& 79.59$\pm$2.57& 19.85$\pm$0.44& 194.64$\pm$1.86 & 106.66$\pm$3.87 & 16.80$\pm$0.50 \\
LightPath (w/o ft) & \underline{129.48$\pm$0.26} & \underline{56.82$\pm$2.58} & \underline{12.71$\pm$1.00} & \underline{186.02$\pm$3.96} & \underline{77.33$\pm$2.59} & \underline{10.41$\pm$0.38} \\
START (w/o ft) & 144.54$\pm$0.90& 79.78$\pm$0.87& 19.72$\pm$0.27& 213.22$\pm$2.19 & 120.74$\pm$2.70 & 20.01$\pm$0.49 \\
\textbf{PTrajM} (w/o ft) & \textbf{104.61$\pm$1.05} & \textbf{50.51$\pm$0.87} & \textbf{11.88$\pm$0.23} & \textbf{155.70$\pm$0.25} & \textbf{71.28$\pm$0.78} & \textbf{10.23$\pm$0.18} \\
\midrule
t2vec & 127.41$\pm$2.68& 64.67$\pm$3.58& 14.01$\pm$0.71& 214.40$\pm$2.05 & 108.80$\pm$2.01 & 16.96$\pm$0.94 \\
Trembr & 124.32$\pm$3.67& 63.42$\pm$0.57& 13.60$\pm$0.28& 209.12$\pm$3.02 & 107.02$\pm$1.39 & 16.40$\pm$0.86 \\
CTLE & 135.21$\pm$14.97& \underline{55.41$\pm$7.17} & \underline{11.18$\pm$1.52} & 207.16$\pm$7.44 & 107.46$\pm$9.04 & 16.25$\pm$2.94 \\
Toast & 171.58$\pm$49.56& 91.66$\pm$57.29 & 18.84$\pm$13.04& 202.99$\pm$36.20 & 102.73$\pm$26.14 & 15.75$\pm$2.24 \\
TrajCL & 132.98$\pm$1.06& 55.78$\pm$0.89 & 11.86$\pm$0.23& 183.74$\pm$2.54 & 73.21$\pm$3.45 & 12.55$\pm$0.45 \\
LightPath & 123.00$\pm$8.85& 58.04$\pm$8.56& 12.83$\pm$1.70& 169.01$\pm$1.94 & 74.08$\pm$3.13 & 10.50$\pm$0.41 \\
START & \underline{121.11$\pm$16.25} & 58.97$\pm$11.59& 13.49$\pm$2.57& \underline{159.89$\pm$4.55} & \underline{72.19$\pm$3.09} & \underline{10.26$\pm$0.35} \\
\textbf{PTrajM} & \textbf{46.37$\pm$2.15} & \textbf{18.08$\pm$2.33} & \textbf{4.40$\pm$0.74} & \textbf{86.08$\pm$7.27} & \textbf{30.80$\pm$2.88} & \textbf{4.15$\pm$0.35} \\
\bottomrule
\end{tabular}
\begin{tablenotes}\footnotesize
\item[]{
    \textbf{Bold} denotes the best result, and \underline{underline} denotes the second-best result.
    $\downarrow$ means lower is better.
}
\end{tablenotes}
\end{threeparttable}
\end{table*}

\begin{table*}
\begin{minipage}{0.58\linewidth}
    \caption{Overall performance of methods on similar trajectory search.}
    \label{tab:overall-similar-trajectory-search}
    \resizebox{1.0\linewidth}{!}{
\begin{threeparttable}
\begin{tabular}{c|ccc|ccc}
\toprule
Dataset & \multicolumn{3}{c|}{Chengdu} & \multicolumn{3}{c}{Xian} \\
\midrule
\multirow{2}{*}{\small \diagbox[]{Method}{Metric}} & Acc@1 $\uparrow$ & Acc@5 $\uparrow$ & Mean $\downarrow$ & Acc@1 $\uparrow$ & Acc@5 $\uparrow$ & Mean $\downarrow$ \\
& (\%) & (\%) & Rank & (\%) & (\%) & Rank \\
\midrule
t2vec & 81.45$\pm$0.78& 93.70$\pm$1.84& 3.35$\pm$0.38& 89.47$\pm$3.56 & 97.10$\pm$1.64 & 1.60$\pm$0.34 \\
Trembr & 83.98$\pm$1.15& 89.88$\pm$0.30& 4.66$\pm$1.01& 88.00$\pm$1.35 & 93.00$\pm$0.64 & 3.48$\pm$0.96 \\
CTLE & 53.77$\pm$7.41& 69.20$\pm$4.51& 9.43$\pm$1.59& 41.20$\pm$3.83 & 59.80$\pm$9.83 & 6.05$\pm$1.15 \\
Toast & 53.64$\pm$2.24& 71.60$\pm$2.82 & 5.94$\pm$1.13& 30.60$\pm$5.60 & 64.30$\pm$6.50 & 6.18$\pm$1.04 \\
TrajCL & 95.13$\pm$5.02& \underline{98.88$\pm$1.35} & 1.20$\pm$0.22& 95.63$\pm$1.21 & 99.20$\pm$0.12 & \underline{1.09$\pm$0.02} \\
LightPath & 74.27$\pm$4.76& 86.10$\pm$3.87& 27.27$\pm$3.54& 79.63$\pm$3.24 & 91.70$\pm$3.13 & 13.88$\pm$1.23 \\
START & \underline{96.93$\pm$2.06} & \textbf{99.90$\pm$0.10} & \underline{1.09$\pm$0.04} & \underline{95.93$\pm$3.88} & \underline{99.53$\pm$0.76} & 1.14$\pm$0.20 \\
\textbf{PTrajM} & \textbf{98.07$\pm$0.25} & \textbf{99.90$\pm$0.00} & \textbf{1.04$\pm$0.01} & \textbf{99.77$\pm$0.15} & \textbf{100$\pm$0.00} & \textbf{1.00$\pm$0.00} \\
\bottomrule
\end{tabular}
\begin{tablenotes}\footnotesize
\item[]{
    \textbf{Bold} denotes the best result, and \underline{underline} denotes the second-best result.
    $\uparrow$ means higher is better, and $\downarrow$ means lower is better.
}
\end{tablenotes}
\end{threeparttable}
}
\end{minipage}
\hfill
\begin{minipage}{0.39\linewidth}
    \caption{Efficiency of methods.}
    \label{tab:efficiency}
    \resizebox{1.0\linewidth}{!}{
\begin{threeparttable}
\begin{tabular}{c|ccc}
\toprule
Dataset & \multicolumn{3}{c}{Chengdu / Xian} \\
\midrule
\multirow{2}{*}{\small \diagbox[]{Method}{Metric}} & Model size & Train time & Embed time \\
& (MBytes) & (min/epoch) & (seconds) \\
\midrule
t2vec & \textbf{1.641}/\textbf{1.415} & \textbf{2.783}/\textbf{5.937} & 4.445/\underline{9.705} \\
Trembr & 5.752/5.301 & \underline{3.360}/\underline{6.067} & \underline{3.230}/9.723 \\
CTLE & 3.756/3.756 & 4.533/14.354 & 14.581/33.863 \\
Toast & 4.008/3.557 & 4.400/10.650 & 14.540/33.863 \\
TrajCL & 4.382/3.932 & 7.699/14.567 & 10.253/23.877 \\
LightPath & 12.958/12.507 & 10.250/23.217 & 22.486/46.260 \\
START & 15.928/15.026 & 15.927/37.528 & 28.704/49.894 \\
\textbf{PTrajM} & \underline{3.358}/\underline{3.358} & 9.345/29.715 & 
\textbf{1.161}/\textbf{2.571} \\
\bottomrule
\end{tabular}
\begin{tablenotes}\footnotesize
\item[]{
    \textbf{Bold} denotes the best result, and \underline{underline} denotes the second-best result.
}
\end{tablenotes}
\end{threeparttable}
}
\end{minipage}
\end{table*}

\subsection{Performance Comparison}
\subsubsection{Overall Performance}
Tables~\ref{tab:overall-destination-prediction} to~\ref{tab:overall-similar-trajectory-search} compare the overall performance of different methods on the three downstream tasks introduced in Section~\ref{sec:downstream-tasks}. PTrajM consistently shows superior performance across all tasks.

For the destination prediction and arrival time estimation tasks, methods are either pre-trained with fixed parameters or fine-tuned with task supervision. In both cases, PTrajM outperforms the comparison methods. This demonstrates that PTrajM's pre-training process extracts rich semantic information from trajectories without additional task-specific supervision. Moreover, the design of the Trajectory-Mamba model in PTrajM allows it to achieve superior performance with task supervision. For the similar trajectory search task, the methods are pre-trained, and their output embeddings are used for similarity computation. PTrajM achieves the best performance in this task, further highlighting the effectiveness of its pre-training process.

\subsubsection{Efficiency}
Table~\ref{tab:efficiency} compares the efficiency of different methods on both datasets. In terms of model size and embed time, PTrajM demonstrates high computational efficiency, achieving the same lightweight and embed speed as RNN-based methods like TremBR and t2vec. It is significantly more efficient compared to Transformer-based methods like START and LightPath. Given PTrajM's superior performance in a variety of tasks, it achieves its design goal of semantic-rich trajectory learning with high efficiency.

It is worth noting that PTrajM does not have a particularly short training time. However, since the pre-training process does not add extra burden to the embedding process, where efficiency is more critical in real-world applications, the extra training time can be considered worthwhile due to its effectiveness.

\begin{table}
    \caption{Effectiveness of modules.}
    \label{tab:ablation}
    \begin{threeparttable}
\begin{tabular}{c|cc}
\toprule
\multirow{2}{*}{\diagbox[]{Variant}{Metric}} & RMSE $\downarrow$ & MAE $\downarrow$ \\
& (meters) & (meters) \\
\midrule
w/o mb & 341.13$\pm$1.30 & 273.13$\pm$0.60\\
w/o POI & 387.12$\pm$18.86 & 310.03$\pm$13.99\\
w/o road & 345.35$\pm$4.45 & 276.48$\pm$2.17\\
full &  \textbf{332.06$\pm$7.20} & \textbf{260.38$\pm$6.75} \\
\bottomrule
\end{tabular}
\end{threeparttable}
\end{table}

\subsection{Model Analysis}
We perform analysis on the modules and hyper-parameters of PTrajM on Chengdu dataset, destination prediction task, with the \textit{w/o ft} setting.

\subsubsection{Effectiveness of Modules}
We compare the \textit{full} PTrajM method with the following variants:
\begin{enumerate}[leftmargin=*]
    \item \textit{w/o mb}: replace the movement behavior parameterization with the vanilla input parameterization in Mamba.
    \item \textit{w/o POI}: remove the POI view in the pre-training.
    \item \textit{w/o road}: remove the road view in the pre-training.
\end{enumerate}

Table~\ref{tab:ablation} compares the results. We observe that removing the movement behavior parameterization negatively impacts performance, highlighting the module's effectiveness in extracting movement behavior. Removing either the POI or the road view from the pre-training also leads to worse performance, showing that both contribute to modeling semantic information.

\subsubsection{Effectiveness of Hyper-parameters}
Figure~\ref{fig:hyper-parameter} illustrates the effectiveness of key hyper-parameters. We observe the following:
\begin{enumerate}[leftmargin=*]
    \item The batch size $B$ mainly controls the number of negative trajectories in pre-training, with an optimal value of 128.
    \item The number of layers $L$, state dimension $N$, embed dimension $E$, and model dimension $D$ control the model capacity. $E$ has the most prominent effect since it directly controls the dimension of the final trajectory embeddings. After balancing performance and efficiency, their optimal values are $L=4$, $N=128$, $E=256$, and $D=256$.
    \item The number of heads $H$ determines the complexity of the multi-input SSM in Trajectory-Mamba, with an optimal value of 4.
\end{enumerate}

\section{Conclusion}
We propose PTrajM, a new method for efficient and semantic-rich trajectory learning. First, Trajectory-Mamba is introduced as the learnable model of PTrajM. It parameterizes high-order movement behavior features and integrates them into a trajectory state-space model, enabling PTrajM to effectively and efficiently extract continuous movement behavior. Second, a travel purpose-aware pre-training procedure is proposed to help PTrajM extract travel purposes from trajectories while maintaining its efficiency. Finally, extensive experiments on two real-world vehicle trajectories and three representative tasks demonstrate PTrajM’s effectiveness.



\bibliographystyle{ACM-Reference-Format}
\bibliography{reference}


\begin{thebibliography}{47}


\ifx \showCODEN    \undefined \def \showCODEN     #1{\unskip}     \fi
\ifx \showDOI      \undefined \def \showDOI       #1{#1}\fi
\ifx \showISBNx    \undefined \def \showISBNx     #1{\unskip}     \fi
\ifx \showISBNxiii \undefined \def \showISBNxiii  #1{\unskip}     \fi
\ifx \showISSN     \undefined \def \showISSN      #1{\unskip}     \fi
\ifx \showLCCN     \undefined \def \showLCCN      #1{\unskip}     \fi
\ifx \shownote     \undefined \def \shownote      #1{#1}          \fi
\ifx \showarticletitle \undefined \def \showarticletitle #1{#1}   \fi
\ifx \showURL      \undefined \def \showURL       {\relax}        \fi
\providecommand\bibfield[2]{#2}
\providecommand\bibinfo[2]{#2}
\providecommand\natexlab[1]{#1}
\providecommand\showeprint[2][]{arXiv:#2}

\bibitem[Brown et~al\mbox{.}(2020)]%
        {DBLP:conf/nips/BrownMRSKDNSSAA20}
\bibfield{author}{\bibinfo{person}{Tom~B. Brown}, \bibinfo{person}{Benjamin Mann}, \bibinfo{person}{Nick Ryder}, \bibinfo{person}{Melanie Subbiah}, \bibinfo{person}{Jared Kaplan}, \bibinfo{person}{Prafulla Dhariwal}, \bibinfo{person}{Arvind Neelakantan}, \bibinfo{person}{Pranav Shyam}, \bibinfo{person}{Girish Sastry}, \bibinfo{person}{Amanda Askell}, \bibinfo{person}{Sandhini Agarwal}, \bibinfo{person}{Ariel Herbert{-}Voss}, \bibinfo{person}{Gretchen Krueger}, \bibinfo{person}{Tom Henighan}, \bibinfo{person}{Rewon Child}, \bibinfo{person}{Aditya Ramesh}, \bibinfo{person}{Daniel~M. Ziegler}, \bibinfo{person}{Jeffrey Wu}, \bibinfo{person}{Clemens Winter}, \bibinfo{person}{Christopher Hesse}, \bibinfo{person}{Mark Chen}, \bibinfo{person}{Eric Sigler}, \bibinfo{person}{Mateusz Litwin}, \bibinfo{person}{Scott Gray}, \bibinfo{person}{Benjamin Chess}, \bibinfo{person}{Jack Clark}, \bibinfo{person}{Christopher Berner}, \bibinfo{person}{Sam McCandlish}, \bibinfo{person}{Alec Radford}, \bibinfo{person}{Ilya Sutskever},
  {and} \bibinfo{person}{Dario Amodei}.} \bibinfo{year}{2020}\natexlab{}.
\newblock \showarticletitle{Language Models are Few-Shot Learners}. In \bibinfo{booktitle}{\emph{NeurIPS}}.
\newblock


\bibitem[Chang et~al\mbox{.}(2023)]%
        {DBLP:conf/icde/Chang0LT23}
\bibfield{author}{\bibinfo{person}{Yanchuan Chang}, \bibinfo{person}{Jianzhong Qi}, \bibinfo{person}{Yuxuan Liang}, {and} \bibinfo{person}{Egemen Tanin}.} \bibinfo{year}{2023}\natexlab{}.
\newblock \showarticletitle{Contrastive Trajectory Similarity Learning with Dual-Feature Attention}. In \bibinfo{booktitle}{\emph{ICDE}}. \bibinfo{pages}{2933--2945}.
\newblock


\bibitem[Chao et~al\mbox{.}(2020)]%
        {DBLP:conf/adc/ChaoXH020}
\bibfield{author}{\bibinfo{person}{Pingfu Chao}, \bibinfo{person}{Yehong Xu}, \bibinfo{person}{Wen Hua}, {and} \bibinfo{person}{Xiaofang Zhou}.} \bibinfo{year}{2020}\natexlab{}.
\newblock \showarticletitle{A Survey on Map-Matching Algorithms}. In \bibinfo{booktitle}{\emph{ADC}}, Vol.~\bibinfo{volume}{12008}. \bibinfo{pages}{121--133}.
\newblock


\bibitem[Chen et~al\mbox{.}(2020)]%
        {DBLP:conf/icml/ChenK0H20}
\bibfield{author}{\bibinfo{person}{Ting Chen}, \bibinfo{person}{Simon Kornblith}, \bibinfo{person}{Mohammad Norouzi}, {and} \bibinfo{person}{Geoffrey~E. Hinton}.} \bibinfo{year}{2020}\natexlab{}.
\newblock \showarticletitle{A Simple Framework for Contrastive Learning of Visual Representations}. In \bibinfo{booktitle}{\emph{ICML}}, Vol.~\bibinfo{volume}{119}. \bibinfo{pages}{1597--1607}.
\newblock


\bibitem[Chen et~al\mbox{.}(2018)]%
        {DBLP:conf/nips/ChenRBD18}
\bibfield{author}{\bibinfo{person}{Tian~Qi Chen}, \bibinfo{person}{Yulia Rubanova}, \bibinfo{person}{Jesse Bettencourt}, {and} \bibinfo{person}{David Duvenaud}.} \bibinfo{year}{2018}\natexlab{}.
\newblock \showarticletitle{Neural Ordinary Differential Equations}. In \bibinfo{booktitle}{\emph{NeurIPS}}. \bibinfo{pages}{6572--6583}.
\newblock


\bibitem[Chen et~al\mbox{.}(2022)]%
        {DBLP:conf/ijcai/ChenLHYJD22}
\bibfield{author}{\bibinfo{person}{Wei Chen}, \bibinfo{person}{Shuzhe Li}, \bibinfo{person}{Chao Huang}, \bibinfo{person}{Yanwei Yu}, \bibinfo{person}{Yongguo Jiang}, {and} \bibinfo{person}{Junyu Dong}.} \bibinfo{year}{2022}\natexlab{}.
\newblock \showarticletitle{Mutual Distillation Learning Network for Trajectory-User Linking}. In \bibinfo{booktitle}{\emph{IJCAI}}. \bibinfo{pages}{1973--1979}.
\newblock


\bibitem[Chen et~al\mbox{.}(2021)]%
        {DBLP:conf/cikm/ChenLCBLLCE21}
\bibfield{author}{\bibinfo{person}{Yile Chen}, \bibinfo{person}{Xiucheng Li}, \bibinfo{person}{Gao Cong}, \bibinfo{person}{Zhifeng Bao}, \bibinfo{person}{Cheng Long}, \bibinfo{person}{Yiding Liu}, \bibinfo{person}{Arun~Kumar Chandran}, {and} \bibinfo{person}{Richard Ellison}.} \bibinfo{year}{2021}\natexlab{}.
\newblock \showarticletitle{Robust Road Network Representation Learning: When Traffic Patterns Meet Traveling Semantics}. In \bibinfo{booktitle}{\emph{CIKM}}. \bibinfo{pages}{211--220}.
\newblock


\bibitem[Chung et~al\mbox{.}(2014)]%
        {DBLP:journals/corr/ChungGCB14}
\bibfield{author}{\bibinfo{person}{Junyoung Chung}, \bibinfo{person}{Caglar Gulcehre}, \bibinfo{person}{KyungHyun Cho}, {and} \bibinfo{person}{Yoshua Bengio}.} \bibinfo{year}{2014}\natexlab{}.
\newblock \showarticletitle{Empirical evaluation of gated recurrent neural networks on sequence modeling}.
\newblock \bibinfo{journal}{\emph{arXiv preprint arXiv:1412.3555}} (\bibinfo{year}{2014}).
\newblock


\bibitem[Dao and Gu(2024)]%
        {DBLP:journals/corr/abs-2405-21060}
\bibfield{author}{\bibinfo{person}{Tri Dao} {and} \bibinfo{person}{Albert Gu}.} \bibinfo{year}{2024}\natexlab{}.
\newblock \showarticletitle{Transformers are SSMs: Generalized models and efficient algorithms through structured state space duality}.
\newblock \bibinfo{journal}{\emph{arXiv preprint arXiv:2405.21060}} (\bibinfo{year}{2024}).
\newblock


\bibitem[Devlin et~al\mbox{.}(2019)]%
        {DBLP:conf/naacl/DevlinCLT19}
\bibfield{author}{\bibinfo{person}{Jacob Devlin}, \bibinfo{person}{Ming{-}Wei Chang}, \bibinfo{person}{Kenton Lee}, {and} \bibinfo{person}{Kristina Toutanova}.} \bibinfo{year}{2019}\natexlab{}.
\newblock \showarticletitle{{BERT:} Pre-training of Deep Bidirectional Transformers for Language Understanding}. In \bibinfo{booktitle}{\emph{NAACL}}. \bibinfo{pages}{4171--4186}.
\newblock


\bibitem[Du et~al\mbox{.}(2022)]%
        {DBLP:conf/acl/DuQLDQY022}
\bibfield{author}{\bibinfo{person}{Zhengxiao Du}, \bibinfo{person}{Yujie Qian}, \bibinfo{person}{Xiao Liu}, \bibinfo{person}{Ming Ding}, \bibinfo{person}{Jiezhong Qiu}, \bibinfo{person}{Zhilin Yang}, {and} \bibinfo{person}{Jie Tang}.} \bibinfo{year}{2022}\natexlab{}.
\newblock \showarticletitle{{GLM:} General Language Model Pretraining with Autoregressive Blank Infilling}. In \bibinfo{booktitle}{\emph{ACL}}. \bibinfo{pages}{320--335}.
\newblock


\bibitem[Fang et~al\mbox{.}(2022)]%
        {DBLP:conf/kdd/Fang0ZHCGJ22}
\bibfield{author}{\bibinfo{person}{Ziquan Fang}, \bibinfo{person}{Yuntao Du}, \bibinfo{person}{Xinjun Zhu}, \bibinfo{person}{Danlei Hu}, \bibinfo{person}{Lu Chen}, \bibinfo{person}{Yunjun Gao}, {and} \bibinfo{person}{Christian~S. Jensen}.} \bibinfo{year}{2022}\natexlab{}.
\newblock \showarticletitle{Spatio-Temporal Trajectory Similarity Learning in Road Networks}. In \bibinfo{booktitle}{\emph{KDD}}. \bibinfo{pages}{347--356}.
\newblock


\bibitem[Feng et~al\mbox{.}({[n.\,d.]})]%
        {DBLP:conf/www/FengLZSMGJ18}
\bibfield{author}{\bibinfo{person}{Jie Feng}, \bibinfo{person}{Yong Li}, \bibinfo{person}{Chao Zhang}, \bibinfo{person}{Funing Sun}, \bibinfo{person}{Fanchao Meng}, \bibinfo{person}{Ang Guo}, {and} \bibinfo{person}{Depeng Jin}.} \bibinfo{year}{[n.\,d.]}\natexlab{}.
\newblock \showarticletitle{DeepMove: Predicting Human Mobility with Attentional Recurrent Networks}. In \bibinfo{booktitle}{\emph{WWW}}. \bibinfo{pages}{1459--1468}.
\newblock


\bibitem[Fu and Lee(2020)]%
        {DBLP:journals/tist/FuL20}
\bibfield{author}{\bibinfo{person}{Tao{-}Yang Fu} {and} \bibinfo{person}{Wang{-}Chien Lee}.} \bibinfo{year}{2020}\natexlab{}.
\newblock \showarticletitle{Trembr: Exploring Road Networks for Trajectory Representation Learning}.
\newblock \bibinfo{journal}{\emph{{ACM} Trans. Intell. Syst. Technol.}} \bibinfo{volume}{11}, \bibinfo{number}{1} (\bibinfo{year}{2020}), \bibinfo{pages}{10:1--10:25}.
\newblock


\bibitem[Gu and Dao(2023)]%
        {DBLP:journals/corr/abs-2312-00752}
\bibfield{author}{\bibinfo{person}{Albert Gu} {and} \bibinfo{person}{Tri Dao}.} \bibinfo{year}{2023}\natexlab{}.
\newblock \showarticletitle{Mamba: Linear-time sequence modeling with selective state spaces}.
\newblock \bibinfo{journal}{\emph{arXiv preprint arXiv:2312.00752}} (\bibinfo{year}{2023}).
\newblock


\bibitem[Han et~al\mbox{.}(2022)]%
        {DBLP:journals/pvldb/HanCMG22}
\bibfield{author}{\bibinfo{person}{Xiaolin Han}, \bibinfo{person}{Reynold Cheng}, \bibinfo{person}{Chenhao Ma}, {and} \bibinfo{person}{Tobias Grubenmann}.} \bibinfo{year}{2022}\natexlab{}.
\newblock \showarticletitle{DeepTEA: Effective and Efficient Online Time-dependent Trajectory Outlier Detection}.
\newblock \bibinfo{journal}{\emph{PVLDB}} \bibinfo{volume}{15}, \bibinfo{number}{7} (\bibinfo{year}{2022}), \bibinfo{pages}{1493--1505}.
\newblock


\bibitem[Hinton and Salakhutdinov(2006)]%
        {hinton2006reducing}
\bibfield{author}{\bibinfo{person}{Geoffrey~E Hinton} {and} \bibinfo{person}{Ruslan~R Salakhutdinov}.} \bibinfo{year}{2006}\natexlab{}.
\newblock \showarticletitle{Reducing the dimensionality of data with neural networks}.
\newblock \bibinfo{journal}{\emph{science}} \bibinfo{volume}{313}, \bibinfo{number}{5786} (\bibinfo{year}{2006}), \bibinfo{pages}{504--507}.
\newblock


\bibitem[Hochreiter and Schmidhuber(1997)]%
        {hochreiter1997long}
\bibfield{author}{\bibinfo{person}{Sepp Hochreiter} {and} \bibinfo{person}{J{\"u}rgen Schmidhuber}.} \bibinfo{year}{1997}\natexlab{}.
\newblock \showarticletitle{Long short-term memory}.
\newblock \bibinfo{journal}{\emph{Neural computation}} \bibinfo{volume}{9}, \bibinfo{number}{8} (\bibinfo{year}{1997}), \bibinfo{pages}{1735--1780}.
\newblock


\bibitem[Hu et~al\mbox{.}(2024)]%
        {DBLP:journals/tkde/HuCFFLG24}
\bibfield{author}{\bibinfo{person}{Danlei Hu}, \bibinfo{person}{Lu Chen}, \bibinfo{person}{Hanxi Fang}, \bibinfo{person}{Ziquan Fang}, \bibinfo{person}{Tianyi Li}, {and} \bibinfo{person}{Yunjun Gao}.} \bibinfo{year}{2024}\natexlab{}.
\newblock \showarticletitle{Spatio-Temporal Trajectory Similarity Measures: {A} Comprehensive Survey and Quantitative Study}.
\newblock \bibinfo{journal}{\emph{{IEEE} Trans. Knowl. Data Eng.}} \bibinfo{volume}{36}, \bibinfo{number}{5} (\bibinfo{year}{2024}), \bibinfo{pages}{2191--2212}.
\newblock


\bibitem[Jiang et~al\mbox{.}(2023)]%
        {DBLP:conf/icde/JiangPRJLW23}
\bibfield{author}{\bibinfo{person}{Jiawei Jiang}, \bibinfo{person}{Dayan Pan}, \bibinfo{person}{Houxing Ren}, \bibinfo{person}{Xiaohan Jiang}, \bibinfo{person}{Chao Li}, {and} \bibinfo{person}{Jingyuan Wang}.} \bibinfo{year}{2023}\natexlab{}.
\newblock \showarticletitle{Self-supervised Trajectory Representation Learning with Temporal Regularities and Travel Semantics}. In \bibinfo{booktitle}{\emph{ICDE}}. \bibinfo{pages}{843--855}.
\newblock


\bibitem[Kidger et~al\mbox{.}(2020)]%
        {DBLP:conf/nips/KidgerMFL20}
\bibfield{author}{\bibinfo{person}{Patrick Kidger}, \bibinfo{person}{James Morrill}, \bibinfo{person}{James Foster}, {and} \bibinfo{person}{Terry~J. Lyons}.} \bibinfo{year}{2020}\natexlab{}.
\newblock \showarticletitle{Neural Controlled Differential Equations for Irregular Time Series}. In \bibinfo{booktitle}{\emph{NeurIPS}}.
\newblock


\bibitem[Kong and Wu(2018)]%
        {DBLP:conf/ijcai/Kong018}
\bibfield{author}{\bibinfo{person}{Dejiang Kong} {and} \bibinfo{person}{Fei Wu}.} \bibinfo{year}{2018}\natexlab{}.
\newblock \showarticletitle{{HST-LSTM:} {A} Hierarchical Spatial-Temporal Long-Short Term Memory Network for Location Prediction}. In \bibinfo{booktitle}{\emph{IJCAI}}. \bibinfo{pages}{2341--2347}.
\newblock


\bibitem[Li et~al\mbox{.}({[n.\,d.]})]%
        {DBLP:conf/icde/LiZCJW18}
\bibfield{author}{\bibinfo{person}{Xiucheng Li}, \bibinfo{person}{Kaiqi Zhao}, \bibinfo{person}{Gao Cong}, \bibinfo{person}{Christian~S. Jensen}, {and} \bibinfo{person}{Wei Wei}.} \bibinfo{year}{[n.\,d.]}\natexlab{}.
\newblock \showarticletitle{Deep Representation Learning for Trajectory Similarity Computation}. In \bibinfo{booktitle}{\emph{ICDE}}. \bibinfo{pages}{617--628}.
\newblock


\bibitem[Liang et~al\mbox{.}(2022)]%
        {DBLP:conf/cikm/LiangOWLCZZZ22}
\bibfield{author}{\bibinfo{person}{Yuxuan Liang}, \bibinfo{person}{Kun Ouyang}, \bibinfo{person}{Yiwei Wang}, \bibinfo{person}{Xu Liu}, \bibinfo{person}{Hongyang Chen}, \bibinfo{person}{Junbo Zhang}, \bibinfo{person}{Yu Zheng}, {and} \bibinfo{person}{Roger Zimmermann}.} \bibinfo{year}{2022}\natexlab{}.
\newblock \showarticletitle{TrajFormer: Efficient Trajectory Classification with Transformers}. In \bibinfo{booktitle}{\emph{CIKM}}. \bibinfo{pages}{1229--1237}.
\newblock


\bibitem[Liang et~al\mbox{.}(2021)]%
        {DBLP:conf/ijcai/LiangOYWTZ21}
\bibfield{author}{\bibinfo{person}{Yuxuan Liang}, \bibinfo{person}{Kun Ouyang}, \bibinfo{person}{Hanshu Yan}, \bibinfo{person}{Yiwei Wang}, \bibinfo{person}{Zekun Tong}, {and} \bibinfo{person}{Roger Zimmermann}.} \bibinfo{year}{2021}\natexlab{}.
\newblock \showarticletitle{Modeling Trajectories with Neural Ordinary Differential Equations}. In \bibinfo{booktitle}{\emph{IJCAI}}. \bibinfo{pages}{1498--1504}.
\newblock


\bibitem[Lin et~al\mbox{.}(2023a)]%
        {10375102}
\bibfield{author}{\bibinfo{person}{Yan Lin}, \bibinfo{person}{Huaiyu Wan}, \bibinfo{person}{Shengnan Guo}, \bibinfo{person}{Jilin Hu}, \bibinfo{person}{Christian~S. Jensen}, {and} \bibinfo{person}{Youfang Lin}.} \bibinfo{year}{2023}\natexlab{a}.
\newblock \showarticletitle{Pre-Training General Trajectory Embeddings With Maximum Multi-View Entropy Coding}.
\newblock \bibinfo{journal}{\emph{{IEEE} Trans. Knowl. Data Eng.}} (\bibinfo{year}{2023}), \bibinfo{pages}{1--15}.
\newblock


\bibitem[Lin et~al\mbox{.}(2021)]%
        {DBLP:conf/aaai/LinW0L21}
\bibfield{author}{\bibinfo{person}{Yan Lin}, \bibinfo{person}{Huaiyu Wan}, \bibinfo{person}{Shengnan Guo}, {and} \bibinfo{person}{Youfang Lin}.} \bibinfo{year}{2021}\natexlab{}.
\newblock \showarticletitle{Pre-training Context and Time Aware Location Embeddings from Spatial-Temporal Trajectories for User Next Location Prediction}. In \bibinfo{booktitle}{\emph{AAAI}}. \bibinfo{pages}{4241--4248}.
\newblock


\bibitem[Lin et~al\mbox{.}(2023b)]%
        {DBLP:journals/pacmmod/LinWHGYLJ23}
\bibfield{author}{\bibinfo{person}{Yan Lin}, \bibinfo{person}{Huaiyu Wan}, \bibinfo{person}{Jilin Hu}, \bibinfo{person}{Shengnan Guo}, \bibinfo{person}{Bin Yang}, \bibinfo{person}{Youfang Lin}, {and} \bibinfo{person}{Christian~S. Jensen}.} \bibinfo{year}{2023}\natexlab{b}.
\newblock \showarticletitle{Origin-Destination Travel Time Oracle for Map-based Services}.
\newblock \bibinfo{journal}{\emph{PACMMOD}} \bibinfo{volume}{1}, \bibinfo{number}{3} (\bibinfo{year}{2023}), \bibinfo{pages}{217:1--217:27}.
\newblock


\bibitem[Liu et~al\mbox{.}(2022)]%
        {DBLP:conf/nips/LiuWLW22}
\bibfield{author}{\bibinfo{person}{Xin Liu}, \bibinfo{person}{Zhongdao Wang}, \bibinfo{person}{Yali Li}, {and} \bibinfo{person}{Shengjin Wang}.} \bibinfo{year}{2022}\natexlab{}.
\newblock \showarticletitle{Self-Supervised Learning via Maximum Entropy Coding}. In \bibinfo{booktitle}{\emph{NeurIPS}}.
\newblock


\bibitem[Liu et~al\mbox{.}(2020)]%
        {DBLP:conf/icde/Liu0CB20}
\bibfield{author}{\bibinfo{person}{Yiding Liu}, \bibinfo{person}{Kaiqi Zhao}, \bibinfo{person}{Gao Cong}, {and} \bibinfo{person}{Zhifeng Bao}.} \bibinfo{year}{2020}\natexlab{}.
\newblock \showarticletitle{Online Anomalous Trajectory Detection with Deep Generative Sequence Modeling}. In \bibinfo{booktitle}{\emph{ICDE}}. \bibinfo{pages}{949--960}.
\newblock


\bibitem[Miao et~al\mbox{.}(2020)]%
        {DBLP:conf/wsdm/MiaoLZW20}
\bibfield{author}{\bibinfo{person}{Congcong Miao}, \bibinfo{person}{Ziyan Luo}, \bibinfo{person}{Fengzhu Zeng}, {and} \bibinfo{person}{Jilong Wang}.} \bibinfo{year}{2020}\natexlab{}.
\newblock \showarticletitle{Predicting Human Mobility via Attentive Convolutional Network}. In \bibinfo{booktitle}{\emph{WSDM}}. \bibinfo{pages}{438--446}.
\newblock


\bibitem[Oord et~al\mbox{.}(2018)]%
        {DBLP:journals/corr/abs-1807-03748}
\bibfield{author}{\bibinfo{person}{Aaron van~den Oord}, \bibinfo{person}{Yazhe Li}, {and} \bibinfo{person}{Oriol Vinyals}.} \bibinfo{year}{2018}\natexlab{}.
\newblock \showarticletitle{Representation learning with contrastive predictive coding}.
\newblock \bibinfo{journal}{\emph{arXiv preprint arXiv:1807.03748}} (\bibinfo{year}{2018}).
\newblock


\bibitem[Paszke et~al\mbox{.}(2019)]%
        {DBLP:conf/nips/PaszkeGMLBCKLGA19}
\bibfield{author}{\bibinfo{person}{Adam Paszke}, \bibinfo{person}{Sam Gross}, \bibinfo{person}{Francisco Massa}, \bibinfo{person}{Adam Lerer}, \bibinfo{person}{James Bradbury}, \bibinfo{person}{Gregory Chanan}, \bibinfo{person}{Trevor Killeen}, \bibinfo{person}{Zeming Lin}, \bibinfo{person}{Natalia Gimelshein}, \bibinfo{person}{Luca Antiga}, \bibinfo{person}{Alban Desmaison}, \bibinfo{person}{Andreas K{\"{o}}pf}, \bibinfo{person}{Edward~Z. Yang}, \bibinfo{person}{Zachary DeVito}, \bibinfo{person}{Martin Raison}, \bibinfo{person}{Alykhan Tejani}, \bibinfo{person}{Sasank Chilamkurthy}, \bibinfo{person}{Benoit Steiner}, \bibinfo{person}{Lu Fang}, \bibinfo{person}{Junjie Bai}, {and} \bibinfo{person}{Soumith Chintala}.} \bibinfo{year}{2019}\natexlab{}.
\newblock \showarticletitle{PyTorch: An Imperative Style, High-Performance Deep Learning Library}. In \bibinfo{booktitle}{\emph{NeurIPS}}. \bibinfo{pages}{8024--8035}.
\newblock


\bibitem[Radford et~al\mbox{.}(2021)]%
        {DBLP:conf/icml/RadfordKHRGASAM21}
\bibfield{author}{\bibinfo{person}{Alec Radford}, \bibinfo{person}{Jong~Wook Kim}, \bibinfo{person}{Chris Hallacy}, \bibinfo{person}{Aditya Ramesh}, \bibinfo{person}{Gabriel Goh}, \bibinfo{person}{Sandhini Agarwal}, \bibinfo{person}{Girish Sastry}, \bibinfo{person}{Amanda Askell}, \bibinfo{person}{Pamela Mishkin}, \bibinfo{person}{Jack Clark}, \bibinfo{person}{Gretchen Krueger}, {and} \bibinfo{person}{Ilya Sutskever}.} \bibinfo{year}{2021}\natexlab{}.
\newblock \showarticletitle{Learning Transferable Visual Models From Natural Language Supervision}. In \bibinfo{booktitle}{\emph{ICML}}, Vol.~\bibinfo{volume}{139}. \bibinfo{pages}{8748--8763}.
\newblock


\bibitem[Sang et~al\mbox{.}(2023)]%
        {DBLP:journals/www/SangXCZ23}
\bibfield{author}{\bibinfo{person}{Yu Sang}, \bibinfo{person}{Zhenping Xie}, \bibinfo{person}{Wei Chen}, {and} \bibinfo{person}{Lei Zhao}.} \bibinfo{year}{2023}\natexlab{}.
\newblock \showarticletitle{{TULRN:} Trajectory user linking on road networks}.
\newblock \bibinfo{journal}{\emph{WWW}} \bibinfo{volume}{26}, \bibinfo{number}{4} (\bibinfo{year}{2023}), \bibinfo{pages}{1949--1965}.
\newblock


\bibitem[Tancik et~al\mbox{.}(2020)]%
        {DBLP:conf/nips/TancikSMFRSRBN20}
\bibfield{author}{\bibinfo{person}{Matthew Tancik}, \bibinfo{person}{Pratul~P. Srinivasan}, \bibinfo{person}{Ben Mildenhall}, \bibinfo{person}{Sara Fridovich{-}Keil}, \bibinfo{person}{Nithin Raghavan}, \bibinfo{person}{Utkarsh Singhal}, \bibinfo{person}{Ravi Ramamoorthi}, \bibinfo{person}{Jonathan~T. Barron}, {and} \bibinfo{person}{Ren Ng}.} \bibinfo{year}{2020}\natexlab{}.
\newblock \showarticletitle{Fourier Features Let Networks Learn High Frequency Functions in Low Dimensional Domains}. In \bibinfo{booktitle}{\emph{NeurIPS}}.
\newblock


\bibitem[Vaswani et~al\mbox{.}(2017)]%
        {DBLP:conf/nips/VaswaniSPUJGKP17}
\bibfield{author}{\bibinfo{person}{Ashish Vaswani}, \bibinfo{person}{Noam Shazeer}, \bibinfo{person}{Niki Parmar}, \bibinfo{person}{Jakob Uszkoreit}, \bibinfo{person}{Llion Jones}, \bibinfo{person}{Aidan~N. Gomez}, \bibinfo{person}{Lukasz Kaiser}, {and} \bibinfo{person}{Illia Polosukhin}.} \bibinfo{year}{2017}\natexlab{}.
\newblock \showarticletitle{Attention is All you Need}. In \bibinfo{booktitle}{\emph{NeurIPS}}. \bibinfo{pages}{5998--6008}.
\newblock


\bibitem[Wu et~al\mbox{.}(2017)]%
        {DBLP:conf/ijcai/WuCSZW17}
\bibfield{author}{\bibinfo{person}{Hao Wu}, \bibinfo{person}{Ziyang Chen}, \bibinfo{person}{Weiwei Sun}, \bibinfo{person}{Baihua Zheng}, {and} \bibinfo{person}{Wei Wang}.} \bibinfo{year}{2017}\natexlab{}.
\newblock \showarticletitle{Modeling Trajectories with Recurrent Neural Networks}. In \bibinfo{booktitle}{\emph{IJCAI}}. \bibinfo{pages}{3083--3090}.
\newblock


\bibitem[Yan et~al\mbox{.}(2023)]%
        {DBLP:journals/www/YanZSYD23}
\bibfield{author}{\bibinfo{person}{Bingqi Yan}, \bibinfo{person}{Geng Zhao}, \bibinfo{person}{Lexue Song}, \bibinfo{person}{Yanwei Yu}, {and} \bibinfo{person}{Junyu Dong}.} \bibinfo{year}{2023}\natexlab{}.
\newblock \showarticletitle{PreCLN: Pretrained-based contrastive learning network for vehicle trajectory prediction}.
\newblock \bibinfo{journal}{\emph{WWW}} \bibinfo{volume}{26}, \bibinfo{number}{4} (\bibinfo{year}{2023}), \bibinfo{pages}{1853--1875}.
\newblock


\bibitem[Yang et~al\mbox{.}(2023)]%
        {DBLP:conf/kdd/YangHGYJ23}
\bibfield{author}{\bibinfo{person}{Sean~Bin Yang}, \bibinfo{person}{Jilin Hu}, \bibinfo{person}{Chenjuan Guo}, \bibinfo{person}{Bin Yang}, {and} \bibinfo{person}{Christian~S. Jensen}.} \bibinfo{year}{2023}\natexlab{}.
\newblock \showarticletitle{LightPath: Lightweight and Scalable Path Representation Learning}. In \bibinfo{booktitle}{\emph{KDD}}. \bibinfo{pages}{2999--3010}.
\newblock


\bibitem[Yao et~al\mbox{.}(2019)]%
        {DBLP:conf/icde/YaoCZB19}
\bibfield{author}{\bibinfo{person}{Di Yao}, \bibinfo{person}{Gao Cong}, \bibinfo{person}{Chao Zhang}, {and} \bibinfo{person}{Jingping Bi}.} \bibinfo{year}{2019}\natexlab{}.
\newblock \showarticletitle{Computing Trajectory Similarity in Linear Time: {A} Generic Seed-Guided Neural Metric Learning Approach}. In \bibinfo{booktitle}{\emph{ICDE}}. \bibinfo{pages}{1358--1369}.
\newblock


\bibitem[Yao et~al\mbox{.}(2022)]%
        {DBLP:conf/kdd/YaoHDCHB22}
\bibfield{author}{\bibinfo{person}{Di Yao}, \bibinfo{person}{Haonan Hu}, \bibinfo{person}{Lun Du}, \bibinfo{person}{Gao Cong}, \bibinfo{person}{Shi Han}, {and} \bibinfo{person}{Jingping Bi}.} \bibinfo{year}{2022}\natexlab{}.
\newblock \showarticletitle{TrajGAT: {A} Graph-based Long-term Dependency Modeling Approach for Trajectory Similarity Computation}. In \bibinfo{booktitle}{\emph{KDD}}. \bibinfo{pages}{2275--2285}.
\newblock


\bibitem[Yao et~al\mbox{.}(2017)]%
        {DBLP:conf/ijcnn/YaoZZHB17}
\bibfield{author}{\bibinfo{person}{Di Yao}, \bibinfo{person}{Chao Zhang}, \bibinfo{person}{Zhihua Zhu}, \bibinfo{person}{Jian{-}Hui Huang}, {and} \bibinfo{person}{Jingping Bi}.} \bibinfo{year}{2017}\natexlab{}.
\newblock \showarticletitle{Trajectory clustering via deep representation learning}. In \bibinfo{booktitle}{\emph{IJCNN}}. \bibinfo{pages}{3880--3887}.
\newblock


\bibitem[Yuan et~al\mbox{.}(2020)]%
        {DBLP:conf/sigmod/Yuan0BF20}
\bibfield{author}{\bibinfo{person}{Haitao Yuan}, \bibinfo{person}{Guoliang Li}, \bibinfo{person}{Zhifeng Bao}, {and} \bibinfo{person}{Ling Feng}.} \bibinfo{year}{2020}\natexlab{}.
\newblock \showarticletitle{Effective Travel Time Estimation: When Historical Trajectories over Road Networks Matter}. In \bibinfo{booktitle}{\emph{SIGMOD}}. \bibinfo{pages}{2135--2149}.
\newblock


\bibitem[Zhou et~al\mbox{.}(2021)]%
        {DBLP:journals/kbs/ZhouDGWZ21}
\bibfield{author}{\bibinfo{person}{Fan Zhou}, \bibinfo{person}{Yurou Dai}, \bibinfo{person}{Qiang Gao}, \bibinfo{person}{Pengyu Wang}, {and} \bibinfo{person}{Ting Zhong}.} \bibinfo{year}{2021}\natexlab{}.
\newblock \showarticletitle{Self-supervised human mobility learning for next location prediction and trajectory classification}.
\newblock \bibinfo{journal}{\emph{Knowl. Based Syst.}}  \bibinfo{volume}{228} (\bibinfo{year}{2021}), \bibinfo{pages}{107214}.
\newblock


\bibitem[Zhou et~al\mbox{.}(2023)]%
        {DBLP:journals/www/ZhouHYCZ23}
\bibfield{author}{\bibinfo{person}{Silin Zhou}, \bibinfo{person}{Peng Han}, \bibinfo{person}{Di Yao}, \bibinfo{person}{Lisi Chen}, {and} \bibinfo{person}{Xiangliang Zhang}.} \bibinfo{year}{2023}\natexlab{}.
\newblock \showarticletitle{Spatial-temporal fusion graph framework for trajectory similarity computation}.
\newblock \bibinfo{journal}{\emph{WWW}} \bibinfo{volume}{26}, \bibinfo{number}{4} (\bibinfo{year}{2023}), \bibinfo{pages}{1501--1523}.
\newblock


\bibitem[Zhou et~al\mbox{.}(2024)]%
        {DBLP:journals/corr/abs-2405-12459}
\bibfield{author}{\bibinfo{person}{Zeyu Zhou}, \bibinfo{person}{Yan Lin}, \bibinfo{person}{Haomin Wen}, \bibinfo{person}{Shengnan Guo}, \bibinfo{person}{Jilin Hu}, \bibinfo{person}{Youfang Lin}, {and} \bibinfo{person}{Huaiyu Wan}.} \bibinfo{year}{2024}\natexlab{}.
\newblock \showarticletitle{PLM4Traj: Cognizing Movement Patterns and Travel Purposes from Trajectories with Pre-trained Language Models}.
\newblock \bibinfo{journal}{\emph{arXiv preprint arXiv:2405.12459}} (\bibinfo{year}{2024}).
\newblock


\end{thebibliography}

\end{document}